# A Confidence-Aware Multimodal Fusion Framework for Industrial Human-Robot Collaboration

Xinyu Liu, Qiqi Dong, Boya Jia, Yi Zhang, Binbin Lian

***Abstract*—A confidence-aware multimodal fusion framework (CAMF) is proposed to realize reliable human intention prediction for industrial human-robot collaboration. This framework fuses four heterogeneous modalities including object 6D pose, gaze, skeletal motion and IMU-based hand motion. It embeds a confidence-trend-driven dynamic fusion mechanism into BiLSTM to adaptively balance bidirectional temporal features according to real-time modality reliability. A confidence-guided balanced learning strategy combined with a confidence freezing mechanism is further adopted to adjust network gradients dynamically, suppress noise from low-quality modalities and mitigate cross-modal learning bias. A physical platform based on the UR3 collaborative robot is built for experimental validation. Comparative results show that the proposed method reaches an intention recognition accuracy of 91.86% and outperforms existing multimodal fusion approaches in overall performance and stability. It also maintains satisfactory accuracy under low light and partial occlusion interference. In practical assembly tasks, the framework enables proactive and stable human-robot cooperation with strong environmental adaptability.**

***Note to Practitioners*— In the field of industrial human-robot collaboration, accurate human intention understanding serves as a core prerequisite for realizing safe, efficient and seamless collaborative assembly tasks in practical workshops. Single-modal perception is vulnerable to varying lighting conditions, human occlusions and inherent sensor noise in complex working environments. Current multimodal fusion methods mostly adopt static fusion strategies, which fail to dynamically evaluate real-time modality reliability and commonly suffer from imbalanced learning across different data modalities. They also lack the capability of proactive human intention prediction, greatly limiting their practical deployment in high-precision and fast-paced industrial tasks. This paper presents a confidence-aware multimodal fusion framework (CAMF) to tackle the above challenging issues. The system integrates depth camera and 9-axis IMU to capture multi-source human behavioral data, and leverages a series of confidence-related mechanisms to adaptively optimize feature fusion and model training processes. The proposed algorithm is fully compatible with mainstream sensors and robotic systems built upon ROS2, and it can be easily deployed and applied to a wide range of industrial assembly scenarios. Experimental test results prove that this method possesses strong anti-interference capability and stable performance under complex on-site working conditions. Future work will extend this method to more sophisticated human-robot collaboration tasks.**

## I. INTRODUCTION

THE human-robot collaboration (HRC) technology has been widely applied in industries such as aerospace, heavy equipment manufacturing, and construction [1], [2], [3]. This technology acts as a core pillar of Industry 4.0 and intelligent manufacturing [4]. However, current industrial HRC systems lack the capability of cognitive understanding in complex scenarios, especially in adapting to the uncertain factors introduced by human operators [5]. This deficiency seriously hinders the realization of efficient and seamless human-robot collaboration [6], [7]. To achieve flexible and intelligent human-robot collaboration in complex industrial environments, robot systems must be equipped with human-centric deep cognitive capabilities for complex scenarios. These capabilities enable robots to accurately capture the dynamic changes in human behaviors and working environments.

The core challenge in cognitive understanding of complex scenarios lies in the dynamic and ambiguous nature of scene elements, with humans being the most uncertain factor. Cognitive perception in complex industrial settings goes beyond simple environmental sensing; it requires integrating multidimensional human-related information, such as body movements, gaze, applied force, and interactions with surrounding equipment. Unimodal methods cannot fully capture this rich contextual information, whereas multimodal approaches are needed to support effective perception. This limitation degrades the performance of human–robot collaboration (HRC) systems in dynamic and complex assembly tasks [8], [9], [10].

Multimodal perception and fusion have emerged as a promising solution to overcome the limitations of unimodal methods. This method integrates visual, tactile, inertial, and other sensory data to exploit cross-modal complementarity and redundancy, thereby improving the accuracy and robustness of human intent recognition [11], [12]. Liu et al. [13] proposed a machine learning based multimodal fusion architecture for industrial production environments. The system combines speech, hand motion and body motion. It achieves significantly better performance than any single modality and supports robust human robot collaboration in manufacturing. Wang et al. [14] proposed a function block-based multimodal control method for human–robot collaborative assembly. The method uses haptic, gesture and voice commands to achieve programming-free robot control. Similarly, Duan et al. [15] presented a dual arm robot

HRC framework that fuses gesture, speech, human pose, and visual information. They validated its effectiveness and robustness in a self contained crown-blade-rotor assembly system.

Effective intent understanding requires not only data fusion but also uncertainty-aware strategies that account for inter-modal correlations during the fusion process [16]. Two primary approaches have been explored to address uncertainty in multimodal intent understanding. The first focuses on dynamic fusion during inference, where modality-specific confidence is used as a real-time indicator to adaptively adjust fusion weights [17]. Trick et al. [18] proposed a Bayesian Independent Opinion Pool – based multimodal intent recognition method for eldercare, fusing probability outputs from speech, gesture, gaze, and scene objects via product combination to adaptively assess modality reliability, thereby enhancing robustness and accuracy. Jiang et al. [19] proposed a human-in-the-loop multimodal fusion algorithm that dynamically adjusts modality weights using user feedback within an improved DS theory, effective in challenging scenarios such as muted meetings, low light conditions, and service interactions. The second integrates confidence estimation directly into the training process, embedding reliability modeling into the network architecture so that inter-modal dependencies and uncertainty characteristics are learned end-to-end [20]. Han et al. [21] proposed Multimodal Dynamics, a framework that jointly learns feature-level confidence through bilinear pooling and modality-level confidence via gated mechanisms during training, effectively capturing dynamic reliability variations across modalities. Zhao et al. [22] combined an extended Independent Opinion Pool with constrained batch-wise confidence learning to reduce uncertainty in intent recognition from gesture, speech, and gaze, achieving superior performance over conventional methods in kitchen HRC tasks.

To enable proactive human–robot collaboration, intent recognition must extend beyond the perception of current states. A core objective is thus predictive behavior understanding, which anticipates future actions by modeling the spatiotemporal patterns of ongoing activity. Various approaches have been proposed for this purpose, including recurrent models [23], [24], convolutional networks [25], [26], generative models [27], [28], and Transformers [29], [30]. Among them, Long Short-Term Memory (LSTM)[31] are widely used for intent prediction. Through gated mechanisms, LSTM effectively mitigates the vanishing or exploding gradient problems of conventional RNNs and excels at capturing long-term temporal dependencies, making it well suited for modeling sequential human behaviors with strong temporal structure. To enhance prediction accuracy and robustness in complex scenarios, researchers commonly integrate LSTM with complementary architectures to better capture and fuse temporal features. Chellali et al. [32] proposed a multi-variate LSTM for arm movement prediction in human-robot hand clapping, integrating RGB-D sensor-acquired 3D position data of shoulder, elbow and wrist to predict final hand position and contact time via initial movement samples. Liu et al. [33] developed a CNN-LSTM framework for manufacturing HRC motion prediction, where VGG16 extracts visual features and LSTM models temporal evolution, enabling ongoing motion prediction without wearable sensors for tasks like computer disassembly. Ma et al. [34] designed a Bi-LSTM for sEMG-based continuous upper-limb movement estimation, concatenating forward/backward branches to capture historical/future features, exploit global sequential dependencies in sEMG signals, and mitigate unidirectional models' limitation in processing asynchronous biosignals.

Despite advances in multimodal fusion and intention prediction for human-robot collaboration, existing methods remain inadequate in complex industrial settings. Real-world industrial environments often involve strong lighting, metallic reflections, occlusions, and electromagnetic interference, which severely degrade the reliability of multimodal sensor signals. Current systems lack mechanisms to suppress noisy or low-confidence modalities during training and rely on static fusion strategies during inference, failing to adapt to dynamic changes in modality reliability. This limits overall robustness. Moreover, they generally lack the ability to predict human intentions proactively, hindering timely and seamless collaboration in fast-paced, high-precision industrial tasks.

To address key challenges in complex industrial settings, including dynamically varying modality reliability, noise susceptibility during training, and the lack of proactive intention prediction, we propose a confidence-driven multimodal human-robot collaboration framework named Confidence-Aware Multimodal Fusion Framework (CAMF). CAMF deeply fuses five modalities: object 6D pose, gaze, skeletal motion, and IMU-based hand motion. By estimating real-time confidence for each modality, it dynamically regulates optimization during training by temporarily freezing updates of low-confidence modalities to suppress noise, and adaptively fuses temporal context during inference according to confidence trends. This enables robust and accurate intention prediction, with effectiveness and adaptability validated in real-world assembly tasks.

The main contributions of this work are summarized as follows:

1) We propose a confidence-driven multimodal human–robot interaction framework that deeply fuses four modalities: object 6D pose, gaze, skeletal motion, and IMU-based hand motion. Fusion is adaptively guided by real-time confidence estimates from each modality, and the approach is validated for effectiveness and robustness in real-world assembly scenarios.
2) A balanced multimodal learning method was adopted to adaptively control the optimization of each modality by monitoring the discrepancy in their contributions to the learning objective. A confidence-guided dual mechanism was integrated into this method. The mechanism fuses BiLSTM outputs via confidence-weighted aggregation for final prediction, while temporarily freezing network updates of low-confidence modalities to suppress noise interference.
3) A confidence-trend-driven dynamic fusion mechanism was presented within BiLSTM to adaptively balance

forward and backward temporal contexts according to the time-varying reliability of each modality during interaction.

The rest of the paper is organized as follows. The proposed method is detailed in Section II. Experimental results are presented in Section III. The discussion is provided in Section IV, and the conclusion is given in Section V.

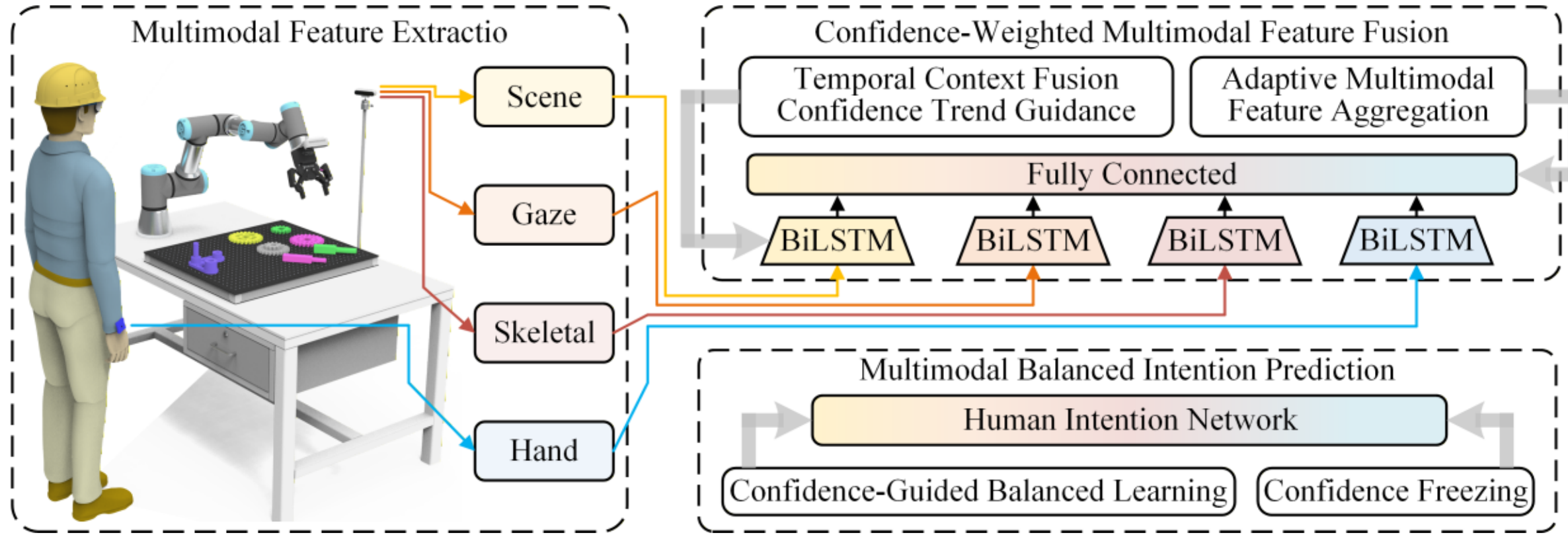

**Fig. 1.** Overall architecture of the Confidence-Aware Multimodal Fusion Framework (CAMF).

## II. Methods

In this section, the overall architecture and working principle of CAMF are introduced. The framework consists of three parts: multimodal feature extraction, feature fusion and intention prediction, as shown in Fig. 1. For multimodal feature extraction, skeletal motion, gaze and object pose are extracted from visual information, and hand motion from IMU. Four types of features from two heterogeneous modalities are processed independently and in parallel to generate recognition results and confidence for each modality. For feature fusion, information of all modalities is integrated with the confidence trend-driven strategy and confidence weighting strategy. For intention prediction, based on the fused features, the final human intention is accurately inferred via the multimodal balanced learning algorithm with the confidence freezing mechanism introduced.

### *A. Multimodal Feature Extraction*

Four human intention-related modalities are extracted in parallel: object 6D pose, gaze, skeletal motion, and hand motion. Object 6D pose, gaze, skeletal motion are from visual data and hand motion from IMU. Each modality is processed independently, with real-time confidence estimation guiding subsequent adaptive fusion and intention prediction.

*1) Scene Recognition:*

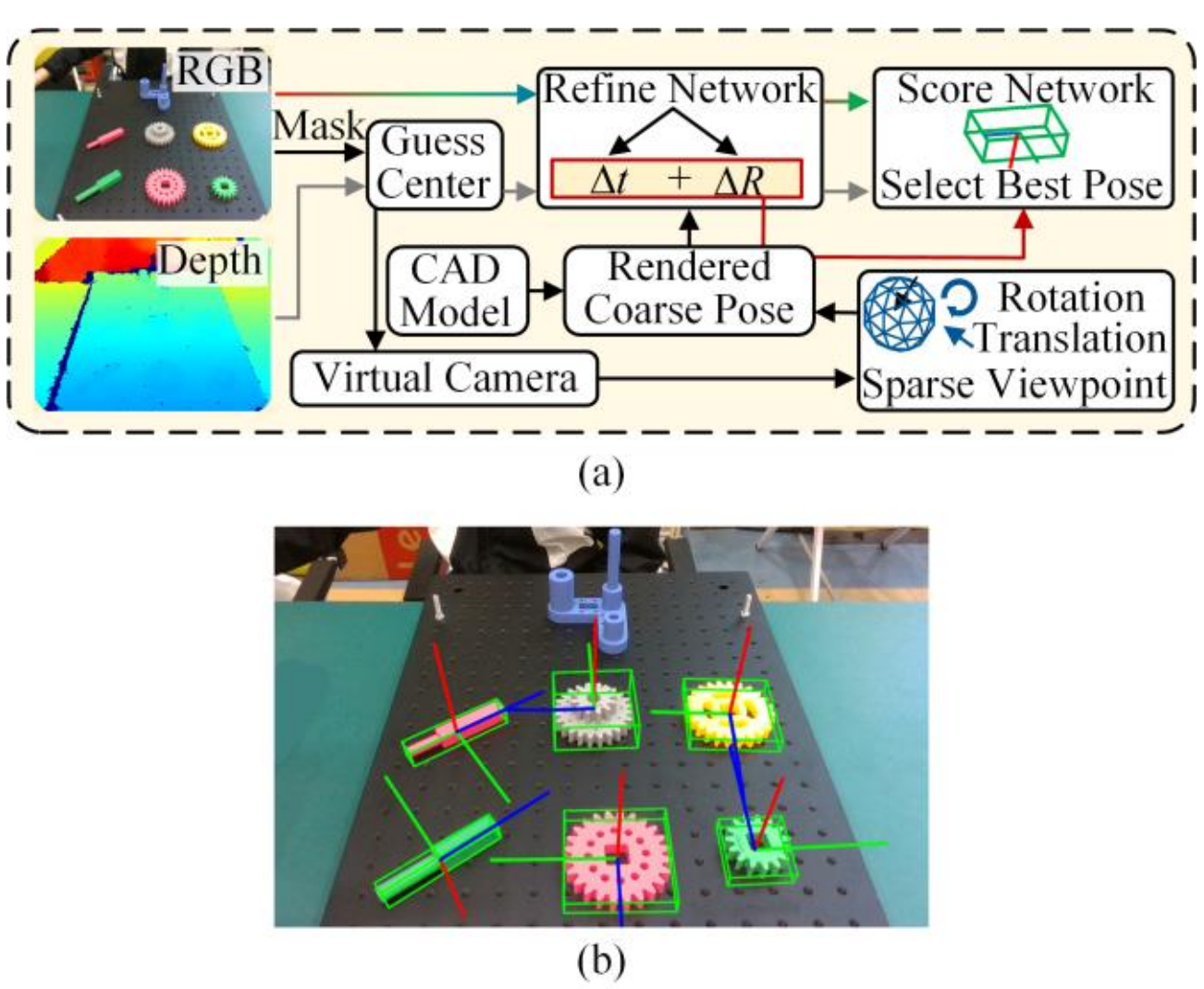

**Fig. 2.** FoundationPose-Based Scene Perception. (a) Scene perception framework. (b) Object pose estimation results.

FoundationPose is employed for 6D object pose estimation and tracking from the input video stream [35]. The overall framework is illustrated in Fig. 2(a). Target object pose and prediction confidence are extracted from RGB-D images, with object mask and CAD model as supplementary inputs. In Pose Initialization, the median 3D point in the mask region is used as the object center. Coarse pose hypotheses are generated by uniform viewpoint sampling on a unit sphere and in-plane rotation discretization. In Pose Refinement, pose hypotheses and cropped image regions are fed into an encoder-residual network that predicts translation and rotation updates. The refined pose is obtained iteratively as follows:

$$\begin{aligned} t^{+} &= t + \Delta t \\ R^{+} &= \Delta R \otimes R \end{aligned} \tag{1}$$

where $t^{+}, R^{+}$ are the refined translation and rotation. are the translation update and rotation update.

In Pose Selection, refined poses are used to render synthetic images from the CAD model. Features from real and synthetic images are concatenated and fed into a scoring module with multi-head self-attention. The network is trained with a pose-conditioned triplet loss as follows:

$$L(i^+,i^-)=\max(S(i^-)-S(i^+)+\varepsilon,0) \tag{2}$$

where $\varepsilon$ is the contrastive margin. $S(i^-),S(i^+)$ are the scores assigned by the scoring module to the positive and negative pose samples.

The pose with the highest score from the scoring module is selected as the final result, as shown in Fig. 2(b). The Scene Recognition confidence is obtained via a Sigmoid nonlinear mapping of this highest score:

$$C_{scene}=\alpha_1\frac{1}{1+\exp(-\alpha_2 P_{scene})} \tag{3}$$

where $P_{scene}$ is the highest score output by the scoring module, and $\alpha_1,\alpha_2$ are weight coefficients.

*2) Gaze Recognition:*

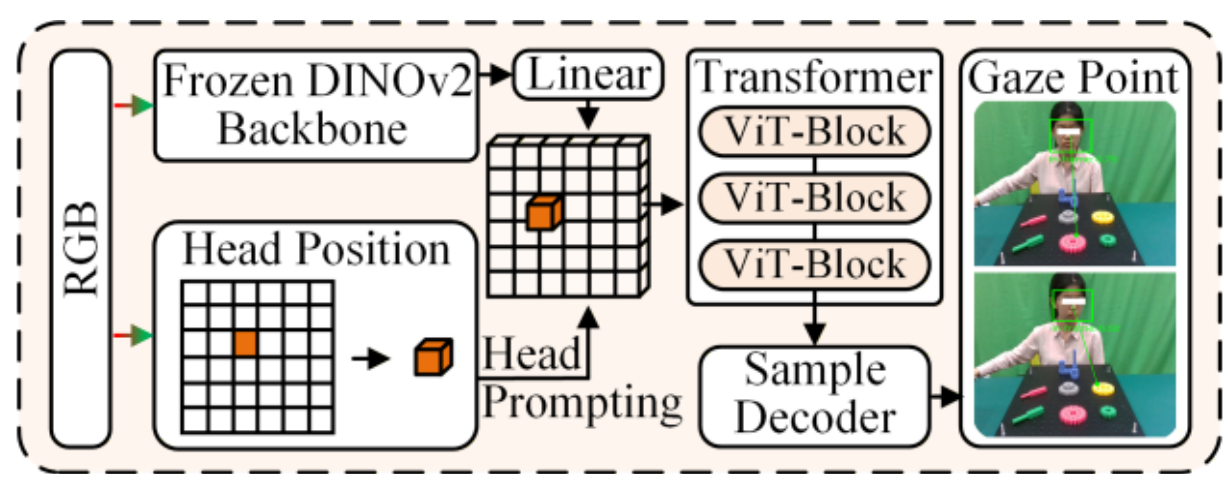

**Fig. 3.** Gaze feature extraction framework.

Gaze Target Estimation via Large-Scale Learned Encoders (Gaze-LLE) is employed to analyze the input video stream, extracting gaze point and confidence [36], is shown in Fig. 3. Gaze-LLE takes scene RGB images and head bounding boxe detected by MediaPipe as inputs. A frozen DINOv2 backbone extracts image features. A Head Position Prompt mechanism encodes head position into learnable embeddings, which modulate the scene feature map additively. A lightweight Transformer decoder processes the modulated features to output a gaze probability heatmap. The gaze point is determined by applying argmax to the heatmap, with the maximum probability as the prediction confidence. The spatial proximity between the gaze point and the target detection frame center is also incorporated to evaluate fixation reliability. The gaze recognition confidence is as follows:

$$\begin{aligned} C_{gaze} &= \beta_1 P_{gaze}+\beta_2 L_{gaze} \\ L_{gaze} &= 1-\tanh(\beta_3\left\|x_{gaze}-x_{object}\right\|) \end{aligned} \tag{4}$$

where $P_{gaze}$ is the maximum probability value from the gaze heatmap. $L_{gaze}$ is the position reliability. $\beta_1,\beta_2,\beta_3$ are weight coefficients. $x_{gaze},x_{object}$ are the gaze point coordinates, and the mean position coordinates of candidate target objects.

*3) Skeletal Motion Recognition:*

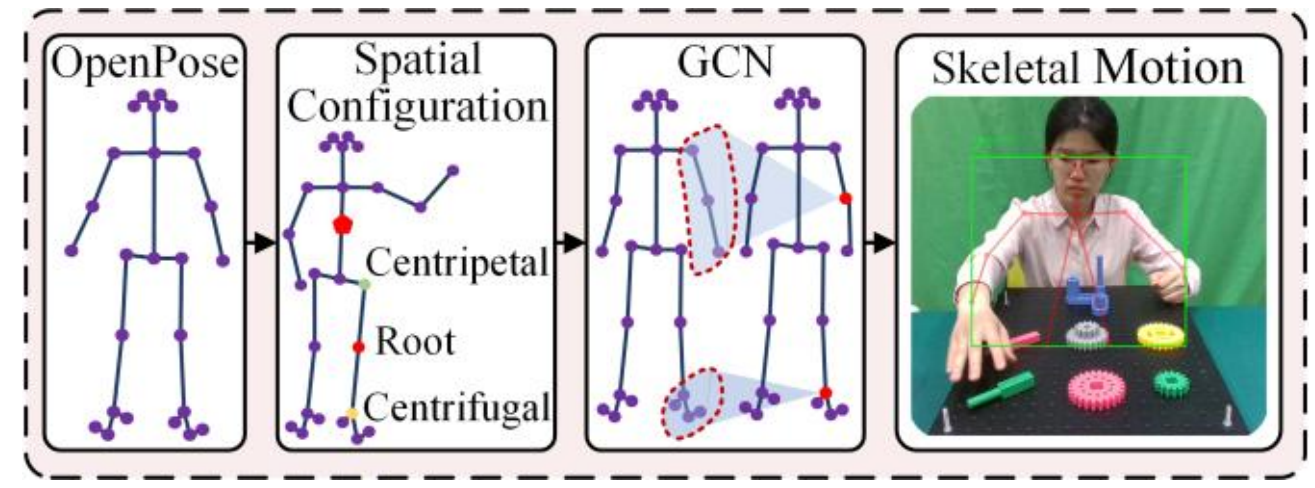

**Fig. 4.** Skeletal motion feature extraction framework.

Spatial Temporal Graph Convolutional Networks (ST-GCN) are adopted to preprocess skeleton data obtained by OpenPose 3D [37], is shown in Fig. 4. The skeleton data is represented as a graph $G=\{V,E\}$, with adjacency matrix $A\in\{0,1\}^{N\times N}$. $V,E$ are the set of joints. and edges connecting adjacent joints. Spatial configuration partitioning divides each node's neighborhood into three subsets. The root node itself. The centripetal group, consisting of neighboring nodes closer to the skeleton's gravity center. The centrifugal group, containing all other nodes. The ST-GCN spatial graph convolution processes the skeleton graph.

$$X_{out}=\sum_k^K M_k\odot\tilde{A}_k X_{in} W_k \tag{5}$$

where $X_{in},X_{out}$ are the input and output features. $M_k$ is the node importance mask. $\tilde{A}_k=\Lambda_k^{-\frac{1}{2}}A_k\Lambda_k^{-\frac{1}{2}}$ is the normalized adjacency matrix. $W_k$ is the learnable weight vector.

Motion rationality confidence evaluates skeleton feature reliability. It incorporates joint spatial consistency and physiological constraint satisfaction:

$$\begin{aligned} C_{skeletal} &= \gamma_1 P_{skeletal}+\gamma_2 L_{skeletal} \\ P_{skeletal} &= \frac{1}{E}\sum_{(i,j)\in E}\left|\left(\left\|x_i-x_j\right\|/L_{ij}\right)-1\right|, x_i,x_j\in X_{in} \\ L_{skeletal} &= \frac{1}{V}\sum_{v\in V}\text{count}(\theta_v\in[\theta_{v,\min},\theta_{v,\max}]) \end{aligned} \tag{6}$$

where $P_{gesture},L_{gesture}$ are joint spatial consistency and physiological constraint satisfaction. $\gamma_1,\gamma_2$ are weight coefficients. $L_{ij}$ is the physiological bone length. $\theta_v$ is the bending angle. $\theta_{v,\min},\theta_{v,\max}$ are the physiological angle ranges.

*4) Hand Motion Recognition:*

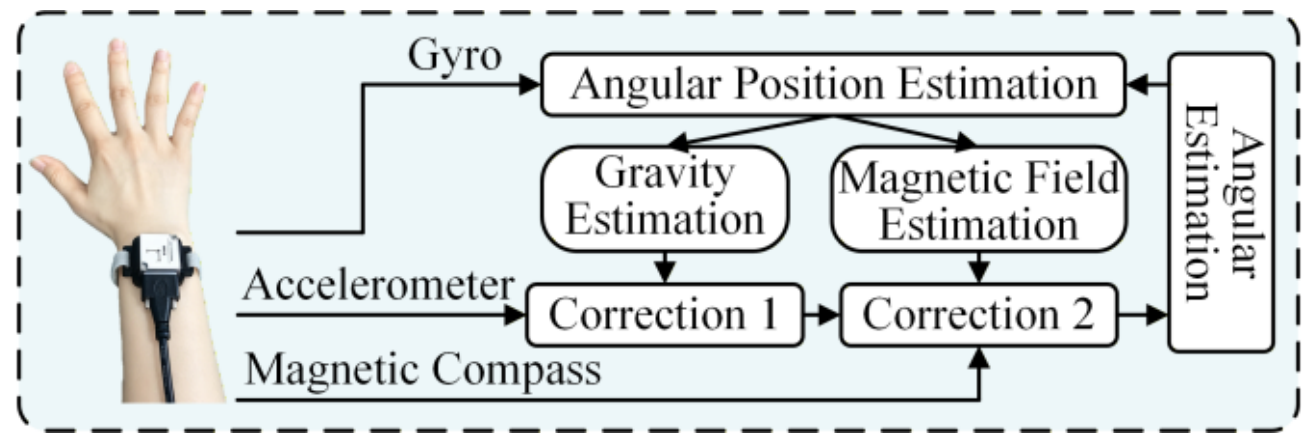

**Fig. 5.** Hand motion feature extraction framework.

A 9-axis inertial measurement unit (IMU) is used to record hand motion, including 3-axis acceleration, 3-axis gyroscope, and 3-axis magnetometer data. A Double-Stage Kalman Filter is employed to estimate the attitude $q_t$, with the corresponding hand motion feature extraction framework illustrated in Fig. 5. The prior attitude $q_t^-$ is predicted by integrating the angular velocity $\omega$ and the attitude estimation at the previous time step.

The first stage corrects pitch and roll using acceleration measurements $z_a$. The theoretical gravity vector is $h_1(q_k^-) = R(q_k^-)[0 \quad 0 \quad |g|]^{\mathrm{T}}$, where $R(q_k^-)$ is the direction cosine matrix and $g$ is the gravitational acceleration. The second stage corrects yaw using magnetometer measurements $z_m$. The theoretical magnetic field is $h_2(q_k^-) = R(q_k^-)[0 \quad 1 \quad 0]^{\mathrm{T}}$. Both stages update the correction via the EKF:

$$
\begin{aligned}
K &= E^- H^{\mathrm{T}} (H E^- H^{\mathrm{T}} + R)^{-1} \\
q &= q^- + K(z - h(q^-))
\end{aligned}
\tag{7}
$$

where $K, H, R$ are the Kalman gain, measurement matrix, and noise covariance matrix. $E^-$ is the prior error covariance.

The hand motion confidence is derived from the prior error covariance and the reliability of the direction.

$$
\begin{aligned}
C_{hand} &= \delta_1 P_{hand} + \delta_2 L_{hand} \\
P_{hand} &= \exp(-\mathrm{tr}(E^-)),\ \ L_{hand} = \max(0, \frac{a \cdot b}{\|a\|\|b\|})
\end{aligned}
\tag{8}
$$

where $P_{hand}, L_{hand}$ are the prediction confidence of the hand pose and the reliability of the direction. $\delta_1, \delta_2$ are weight coefficients. $\mathrm{tr}(P^-)$ is the trace of the matrix. $a$ is the hand motion direction vector obtained by subtracting the gravity component from the accelerometer reading. $b$ is the direction vector from the hand to the mean position coordinates of candidate target objects

*B. Confidence-Weighted Multimodal Feature Fusion*

Human-robot collaboration is essentially a continuous sequential process, where both human intent and environmental state evolve dynamically over time. To effectively capture such temporal dependencies and contextual information, the extracted feature and corresponding confidence value of each modality are fed into an individual BiLSTM network for temporal encoding. The temporally encoded feature representations of each individual modality are then input into fully connected (FC) layers to implement multimodal feature aggregation. During the aggregation process, the weights in the FC layers are adaptively assigned in accordance with the real-time confidence values of each respective modality. This confidence-guided fusion strategy provides a stable and adaptive feature representation for the subsequent dynamic temporal context balancing and final intention inference.

*1) Temporal Context Fusion with Confidence Trend Guidance:* BiLSTM is adopted as the core architecture to capture bidirectional temporal dependencies in human-robot collaboration sequences. It performs temporal encoding on the extracted features of each modality. It overcomes the limitation of unidirectional information propagation in traditional recurrent neural networks. BiLSTM consists of two LSTM units with opposite input directions, namely forward and backward. Its modeling capability fundamentally relies on the internal gating mechanism of LSTM [38].

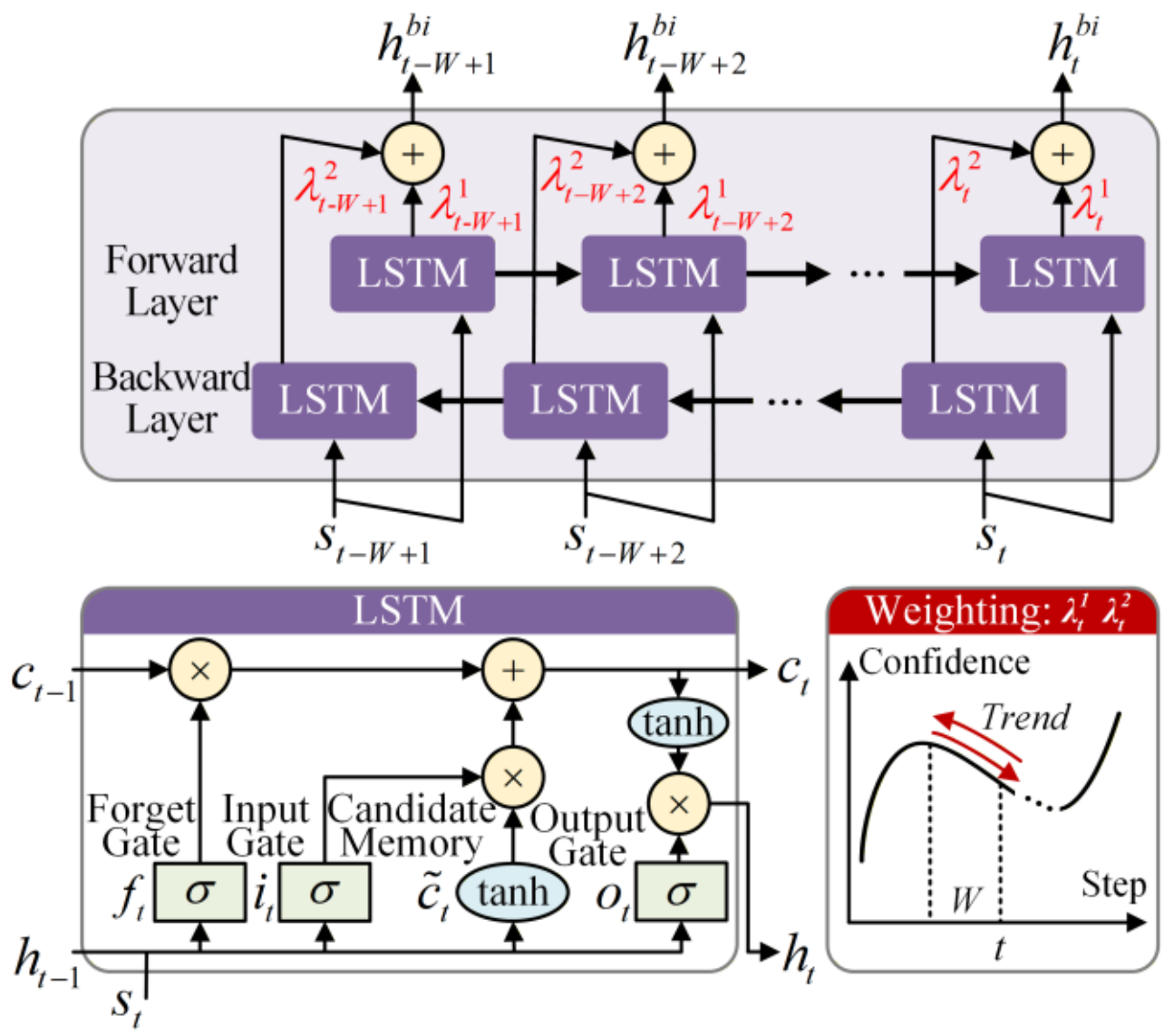

**Fig. 6.** Confidence-trend-driven bidirectional temporal context fusion architecture.

LSTM contains three core gate structures: the forget gate, the input gate, and the output gate. These gates jointly control the retention, update, and output of temporal information.

Forget Gate: This gate determines what information to discard from the cell state.

$$
f_t = \sigma(W_f \cdot [h_{t-1}, s_i] + b_f) \tag{9}
$$

where $f_t$ is the forget gate. $\sigma$ is the sigmoid function. $W_f$ and $b_f$ are the weights and biases. $h_{t-1}$ is the previous hidden state. The hidden state conveys past information to subsequent steps and serves as the current output, offering essential context for sequence modeling.

Input Gate: This gate decides what new information to add to the cell state.

$$
\begin{aligned}
i_t &= \sigma(W_i \cdot [h_{t-1}, s_i] + b_i) \\
\tilde{c}_t &= \tanh(W_c \cdot [h_{t-1}, s_i] + b_c)
\end{aligned}
\tag{10}
$$

where $i_t$ is the input gate. $\tilde{c}_t$ is the candidate memory. $W_i, b, W_c, b_c$ are the respective weights and biases.

$$
c_t = f_t \times c_{t-1} + i_t \times \tilde{c}_t \tag{11}
$$

Output Gate: This gate determines what information to output from the cell state.

$$
o_t = \sigma(W_o \cdot [h_{t-1}, s_i] + b_o) \tag{12}
$$

where $o_t$ is the output gate. $W_o, b_o$ are the weights and biases.

$$
h_t = o_t \times \tanh(c_t) \tag{13}
$$

As shown in Fig. 6, BiLSTM employs two independent LSTM units to model distinct temporal state transitions. The forward unit models the transition from past to current time step $t-1 \rightarrow t$. The backward unit captures the transition from future to current time step $t+1 \rightarrow t$. Within a fixed window of size $W$, the forward LSTM processes the input subsequence $x_{t-W+1}, x_{t-W+2}, ..., x_t$ in chronological order to generate the forward hidden state $\overrightarrow{h_t}$. The backward LSTM processes the same subsequence in reverse order, from $x_t$ to $x_{t-W+1}$, generate the backward hidden state $\overleftarrow{h_t}$. The final bidirectional hidden state at time step $t$ is obtained by fusing the outputs of the two units:

$$h_t^{bi} = \overrightarrow{h_t} + \overleftarrow{h_t} \tag{14}$$

Fixed weighting for bidirectional outputs in BiLSTM cannot adaptively adjust the contribution of forward and backward temporal contexts according to the dynamic characteristics of sequential data. To overcome this limitation, a confidence-trend-driven dynamic weighting strategy is proposed. The dynamically fused bidirectional hidden state is formulated as:

$$h_t^{bi} = \lambda_t^1 \overrightarrow{h_t} + \lambda_t^2 \overleftarrow{h_t} \tag{15}$$

where $\lambda_t^1, \lambda_t^2$ are weight coefficients. They are constructed based on the temporal variation trend of modality confidence. The forward confidence trend $T_t^{for}$ and backward confidence trend $T_t^{back}$ are calculated as follows:

$$\begin{aligned} T_t^{for} &= \frac{1}{W-1} \sum_{k=t-W+1}^{t-1} \frac{C_k - C_{k+1}}{C_k} \\ T_t^{back} &= \frac{1}{W-1} \sum_{k=t-W+2}^{t} \frac{C_k - C_{k-1}}{C_k} \end{aligned} \tag{16}$$

where $C_k$ denotes the confidence value at the *k*-th time step within the sliding window.

Based on these confidence trends, the Softmax function normalizes the indicators to generate adaptive dynamic weights.

$$\begin{aligned} \lambda_t^1 &= \frac{\exp(T_t^{for})}{\exp(T_t^{for}) + \exp(T_t^{back})} \\ \lambda_t^2 &= \frac{\exp(T_t^{back})}{\exp(T_t^{for}) + \exp(T_t^{back})} \end{aligned} \tag{17}$$

When $T_t^{for} > T_t^{back}$, the model assigns a higher weight to the forward LSTM output, relying more on historical evolutionary patterns; when $T_t^{back} > T_t^{for}$, it prioritizes the backward LSTM output, leveraging the more stable future context. This achieves adaptive weighted fusion of bidirectional temporal features.

*2) Adaptive Multimodal Feature Aggregation:* Following temporal encoding via BiLSTM networks, the temporally enhanced multimodal features are aggregated within FC layers, where the contribution of each modality is adaptively weighted according to its confidence value.

LSTM contains three core gate structures: the forget gate, the input gate, and the output gate. These gates jointly control the retention, update, and output of temporal information.

Each sample $x_i = \left(x_i^1, x_i^2, ..., x_i^K\right)$ consists of inputs from $K$ distinct modalities. $y_i \in \{1, 2, \ldots, M\}$ is the corresponding intention category to be predicted, with $M$ representing the total number of intention classes. For each modality, the extracted features $x_i^k$ are fed into a modality-specific BiLSTM encoder $\varphi^k$ with learnable parameters $\theta^k$ to obtain the corresponding hidden state $h_i^k = \varphi^k(\theta^k, x_i^k)$.

FC layer is adopted to fuse the confidence-weighted features of all modalities. $W^{FC}$ and $b^{FC}$ are the weight matrix and bias vector of the fully connected layer.

$$f\left(x_i\right) = W^{FC}\left[C_i^1 h_i^1; C_i^2 h_i^2; ...; C_i^K h_i^K\right] + b^{FC} \tag{18}$$

where $C_i^k$ is the confidence value. This confidence term weights the corresponding modal feature, thereby prioritizing the contributions from more reliable modalities.

## C. Multimodal Balanced Intention Prediction

*1) Confidence-Guided Multimodal Balanced Learning:* High-confidence modalities usually dominate multimodal learning. This suppresses weaker complementary modalities and degrades model generalization. We propose a confidence-guided multimodal balanced learning method, as shown in Fig. 7. It dynamically evaluates modality contributions. It adaptively adjusts gradient scales. It ensures balanced optimization for all modalities.

Network training adopts the gradient descent (GD) approach. In GD optimization framework, the output for intent $m$ is $f(x_i)_m$, and the cross-entropy loss is as follows:

$$L = -\frac{1}{N} \sum_{i=1}^{N} \log \frac{\exp(f(x_i)_{y_i})}{\sum_{m=1}^{M} \exp(f(x_i)_m)} \tag{19}$$

The parameters of $W^{FC}$ and the encoders $\varphi^k$ are updated as:

$$\begin{aligned} W_{t+1}^{FC} &= W_t^{FC} - \eta \nabla_{W^{FC}} L\left(W_t^{FC}\right) \\ &= W_t^{FC} - \eta \frac{1}{N} \sum_{i=1}^{N} \frac{\partial L}{\partial f\left(x_i\right)} \cdot \left[..., C^k h_i^k, ...\right]^{\mathrm{T}} \\ \theta_{t+1}^k &= \theta_t^k - \eta \nabla_{\theta^k} L\left(\theta_t^k\right) \\ &= \theta_t^k - \eta \frac{1}{N} \sum_{i=1}^{N} \frac{\partial L}{\partial f\left(x_i\right)} \cdot C^k \cdot \left(W_t^{FC(k)}\right)^{\mathrm{T}} \cdot \frac{\partial\left(\varphi_t^k\left(\theta^k, x_i^k\right)\right)}{\partial \theta_t^k} \end{aligned} \tag{20}$$

where $\eta$ is the learning rate. $W_t^{FC(k)}$ is the column of $W_t^{FC}$ for modality $k$. The optimization of $W^{FC}, \varphi^k$ is independent of other modalities, except for the loss-related term $\partial L / \partial f\left(x_i\right)$. Thus, encoders cannot adjust parameters based on mutual feedback between modalities.

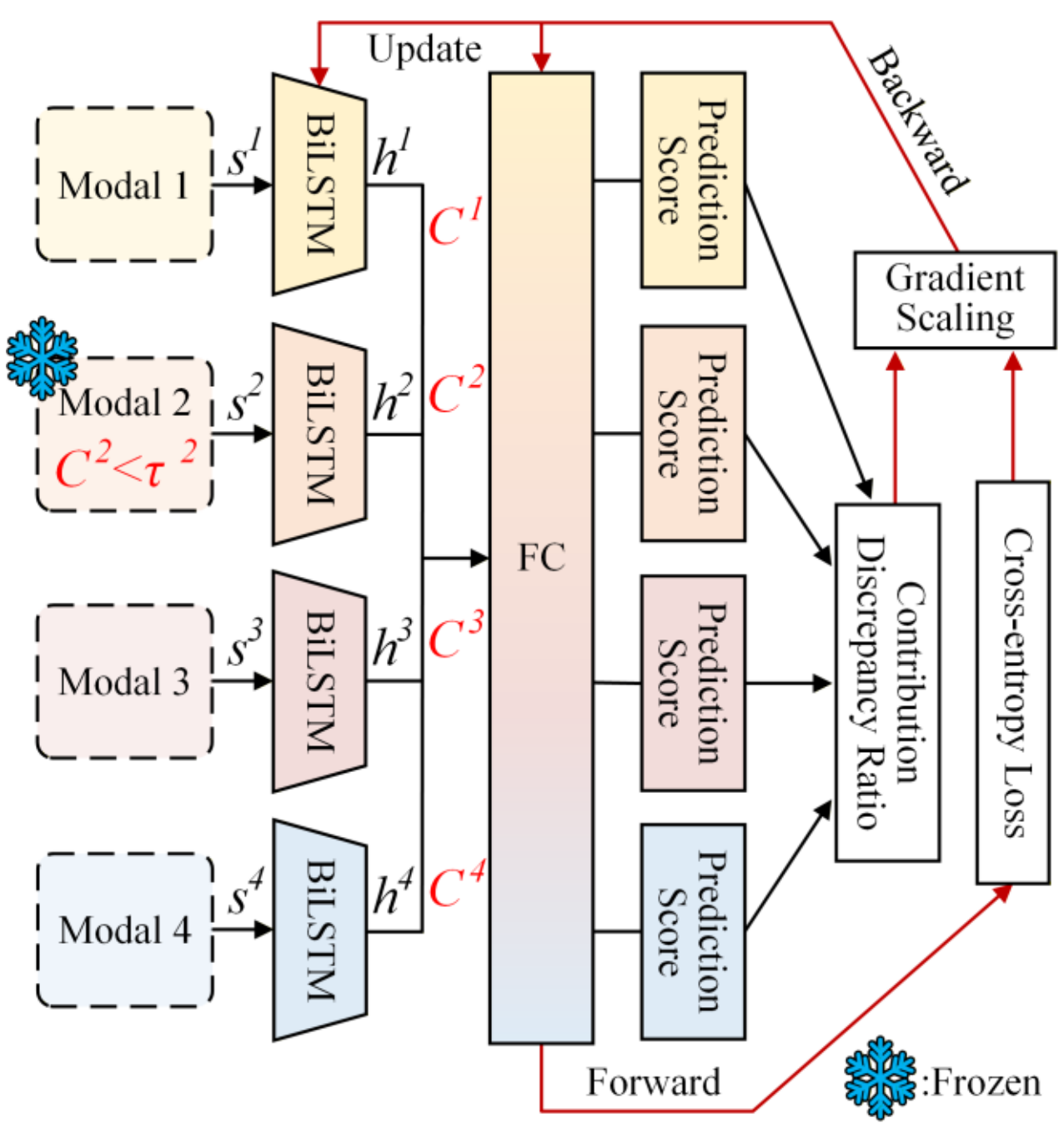

**Fig. 7.** Multimodal Balanced Learning Framework.

In practice, stochastic gradient descent (SGD) is adopted for parameter updates. The update rule is as follows:

$$\theta_{t+1}^{k} = \theta_{t}^{k} - \eta \tilde{g}\left(\theta_{t}^{k}\right) \tag{21}$$

where $\tilde{g}\left(\theta_{t}^{u}\right)$ is an unbiased estimator $\nabla_{\theta^k} L\left(\theta_{t}^{k}\right)$.

For each sample $x_i$, the correct prediction score for each modality is computed as:

$$s_{i}^{k} = \sum_{m=1}^{M} 1_{m=y_i} \cdot \text{softmax}\left(W_{t}^{FC(k)} \cdot C_{i}^{k} h_{i}^{k} + \frac{b^{FC}}{M}\right)_{m} \tag{22}$$

To address the issue that high-confidence modalities usually dominate multimodal learning, a contribution discrepancy ratio is defined to measure the contribution of each modality to the learning objective.

$$\rho_{t}^{k} = \frac{\sum_{i \in B_t} s_{i}^{k}}{\frac{1}{K} \sum_{a \in K} \sum_{i \in B_t} s_{i}^{a}} \tag{23}$$

where $B_t$ is SGD mini-batch of data randomly selected at step $t$.

The gradient scaling coefficient $q_{t}^{k}$ is incorporated to adaptively adjust the balance of modality contributions.

$$q_{t}^{k} = \begin{cases} 1, & \text{if } \rho_{t}^{k} \le 1 \\ 1 - \tanh\left(\kappa \rho_{t}^{k}\right), & \text{if } \rho_{t}^{k} > 1 \end{cases} \tag{24}$$

where $\kappa$ is a scaling weight. When $\rho_{t}^{k} > 1$ , $q_{t}^{k} < 1$ suppressing the gradient updates of that modality.

By incorporating the gradient scaling coefficient into the SGD optimization method, the update rule becomes:

$$\theta_{t+1}^{k} = \theta_{t}^{k} - \eta \cdot q_{t}^{k} \cdot \tilde{g}\left(\theta_{t}^{k}\right) \tag{25}$$

*2) Confidence Freezing Mechanism:* To further enhance the learning stability of the model under extreme noisy conditions, a confidence-gated gradient modulation strategy is introduced. A confidence threshold $\tau$ is set for each modality. If the confidence value of a modality is lower than $\tau$ during the current iteration, its gradient updates are temporarily frozen in that training round. The modulation coefficient is set to $q_{t}^{k} = 0$, which effectively suppresses the interference of low-reliability modalities on model optimization.

## III. Experiment

In this section, comprehensive experiments are carried out to verify the feasibility, robustness and practicality of the proposed CAMF method through quantitative and qualitative analysis. The experiments include hardware and software configuration, dataset construction, ablation studies, comparative analysis of different methods, and human-robot collaborative case tests. Ablation experiments validate the effectiveness of the core confidence-driven fusion strategy and multimodal balanced learning mechanism. Comparative experiments verify that the proposed method outperforms existing mainstream multimodal fusion algorithms in recognition accuracy and stability. Practical case tests further evaluate the environmental adaptability of the model under complex interferences and its reliable performance in real-time human-robot collaborative tasks.

### *A. Experimental Setup*

*1) Experimental Configuration:* This system aims to support multimodal intention recognition and accurate human-robot collaboration, ensuring reliable multimodal data synchronization and meeting real-time inference computational requirements. A complete hardware and software configuration is designed, focusing on compatibility, precision, and practicality in human-robot collaboration scenarios.

For multimodal sensing, an Intel RealSense D435if depth camera captures high-resolution RGB and depth information. It is used to derive the scene, gaze and skeletal motion data. The Hipnuc HI13R3 9-axis IMU accurately captures the user's hand motion and orientation data. The collaborative robotic platform consists of a UR3 6-DOF collaborative manipulator, an Intel RealSense D435if, and a Robotiq 2F-85 adaptive gripper. The UR3 is selected for its high precision and safety. The Robotiq 2F-85 gripper is mounted at the end of the UR3. It can adaptively grasp and stably manipulate the target assembly components, ensuring reliable execution of assembly actions. The Intel RealSense D435if is installed between the robot and the gripper to generate grasping schemes. A Lenovo PX workstation with an NVIDIA A6000 Ada Generation GPU supports real-time multimodal data processing, model inference, and inter-component communication. The software environment is built on Ubuntu 22.04, with Robot Operating System 2 (ROS2) as the core middleware for communication among sensors, workstation, and robot. Time synchronization is implemented across all sensor streams to ensure temporal alignment of multimodal data.

The workpieces used in the experiments are 3D-printed from CAD files provided by the Siemens Robot Learning Challenge,

including baseplate, gear0, gear1, gear2, gear3, shaft0, and shaft1.These components form a standard target assembly task, which is used to validate the effectiveness of the proposed system in intention prediction accuracy and robotic collaboration performance. In the experimental setup, the UR3 is placed beside a worktable, and all assembly components are arranged on the table surface. The D435if covers the entire workspace, and the user wears the IMU to track motion intentions. All sensor data are transmitted to the Lenovo PX workstation for multimodal fusion and intention inference. The generated control commands are sent to the UR3 via ROS2 to execute corresponding actions.

*2) Robot system:* The core function of the UR3 is to perform collaborative operations based on the inferred user intention. The target grasping and delivery process is illustrated in Fig. 8. Upon receiving the user intention command, the system uses the D435if to capture the point cloud of assembly components on the worktable ①, identifies and selects the target workpiece via YOLOv11 with a custom-built dataset ②, extracts the corresponding target mask ③, and employs GraspNet to generate a grasping scheme for the target workpiece ④. Combined with the robotic arm's workspace constraints, the optimal feasible grasping pose is determined, and the gripper is controlled to accomplish stable grasping.

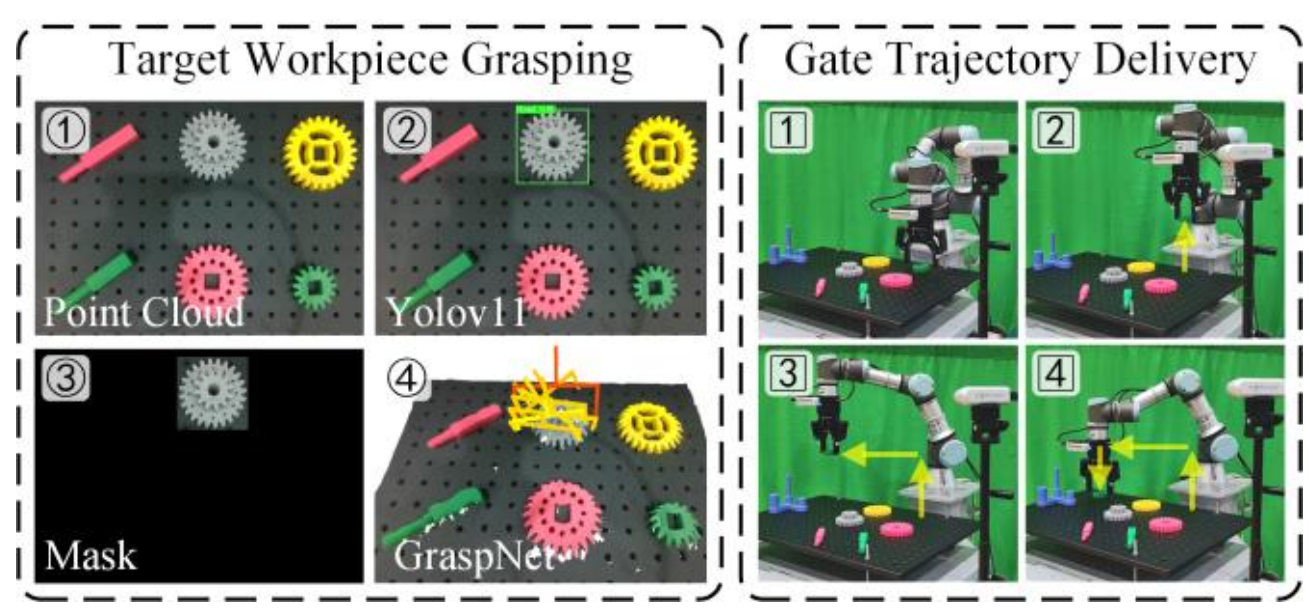

**Fig. 8.** Robotic grasping and delivery process.

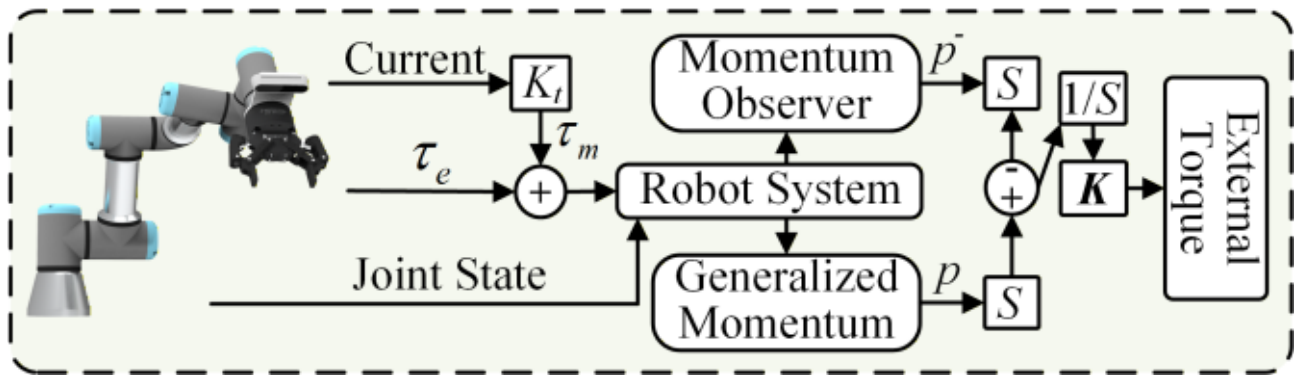

**Fig. 9.** External torque perception framework.

After successfully grasping the target workpiece, the UR3 moves along a safe and efficient gate trajectory [1] [2] [3] [4]. The gate trajectory is designed to avoid collisions with the worktable, surrounding assembly components, and the user. To further guarantee collaborative safety during motion execution, the robot adopts an external torque perception strategy as an auxiliary safety measure, as shown in Fig. 9.

External torque information is estimated and mapped from the joint torque obtained by the momentum observer via the robot Jacobian matrix. The robot dynamic model derived via the Newton-Euler method serves as the foundation for external torque estimation:

$$M(q)\ddot{q} + C(q,\dot{q})\dot{q} + g(q) = \tau_m + \tau_e \quad (26)$$

where $q, \dot{q}, \ddot{q}$ are the joint position, velocity, and acceleration. $M(q)$ is the symmetric positive definite inertia matrix. $C(q,\dot{q})$ is the centrifugal and Coriolis matrix. $g(q)$ is the gravity vector. $\tau_m = K_t I$ is the motor torque, $K_t$ is torque–current constant and $I$ is joint motor current. $\tau_e$ is the total external torque.

Design the first-order momentum observer. Continuously compare the actual generalized momentum $Q = M(q)\dot{q}$ with the observer-estimated momentum $Q^-$, and use a positive-definite gain matrix $K$ to construct a feedback term that estimates the true external disturbance torque.

$$\begin{gathered} \dot{Q}^- = C^T(q,\dot{q})\dot{q} - g(q) - r \\ r = K\left[\int_0^T \left(Q + \tau_e\right) dt - Q\right] \end{gathered} \quad (27)$$

By monitoring external contact torque in real time, the robot can promptly trigger collision avoidance or emergency stop responses when unexpected human-robot contact occurs during trajectory execution. The UR3 then transports the target workpiece along the planned trajectory to the user's operation space. The delivery position enables the user to perform subsequent assembly operations. By completing the entire pipeline from intention recognition, target grasping to workpiece delivery, the robotic arm effectively achieves accurate, safe, and efficient human-robot collaborative assembly.

*3) Dataset Construction:* To evaluate our proposed method, we invited five test participants to engage in the experiment, conducting ten trials for each component, as shown in Fig. 10(a). Prior to formal data acquisition, they engaged in a training session to familiarize themselves with the techniques required for operating the system. The data acquisition procedure consists of three sequential stages, as shown in Fig. 10(b). In S1, each test participant keeps in an initial standby state while waiting for the workpiece-picking instruction. In S2, upon receiving the instruction, the participant starts motion with synchronized sensor data acquisition, approaches the target workpiece via coordinated gaze, body and hand movement, and adjusts trajectory adaptively. In S3, the test participant completes the workpiece-picking action naturally. All test participants perform the procedure naturally without extra intervention, consistent with real-world human operation patterns. Data from each participant are randomly divided into training and test sets at an 8:2 ratio, with 80% for model training and 20% for performance evaluation.

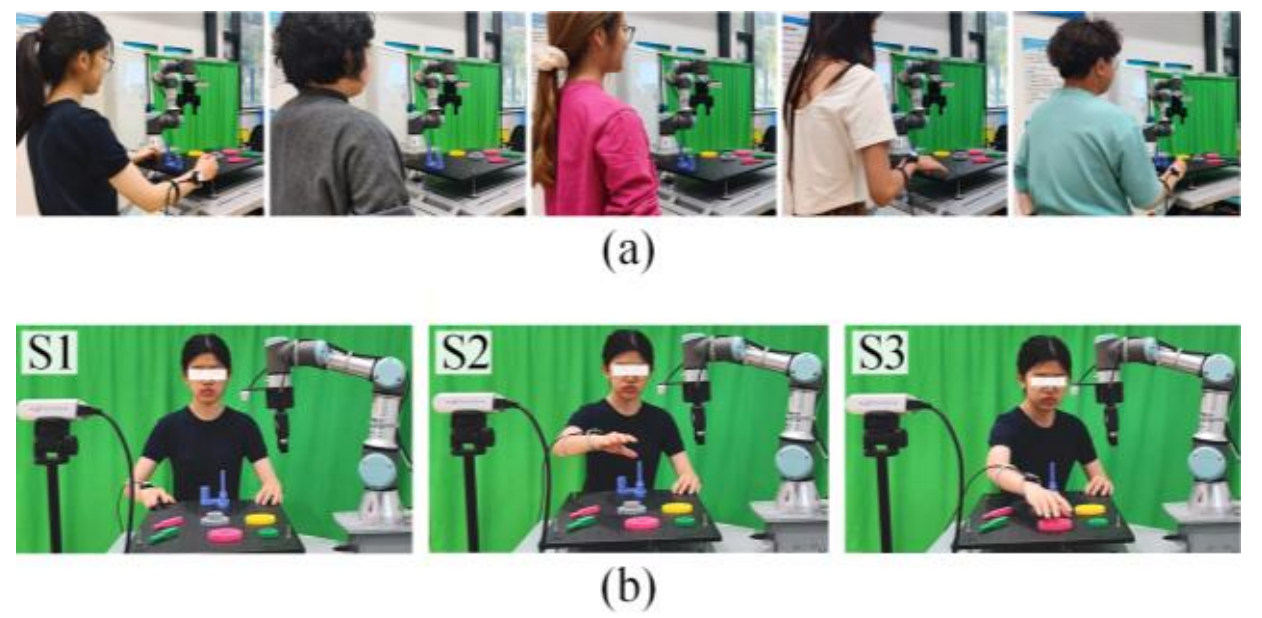

**Fig. 10.** Experimental Setup for Dataset Construction. (a) Test participants. (b) Data acquisition procedure.

*B. Ablation Studies*

*1) Modal Quantity Comparison Results:* This subsection investigates the performance of human intention recognition across different modality combinations. The confusion matrices in Fig. 11 reveal that multimodal approaches achieve far better results than unimodal ones. Scene and gaze modalities exert limited effects on recognition performance. We further conduct quantitative evaluation adopting accuracy, F1 score and Kappa coefficient as evaluation metrics to compare the recognition capability of different modality combinations. The corresponding experimental results are summarized in Table I.

**Fig. 11.** Confusion Matrices of Different Modality Combinations.

TABLE I
THE RESULTS OF DIFFERENT MODALITY COMBINATIONS

| Modality | Accuracy | F1 score | Kappa |
|---|---|---|---|
| Scene | 37.99% | 31.78% | 25.12% |
| Gaze | 43.49% | 43.54% | 32.16% |
| Skeletal | 75.51% | 76.05% | 70.47% |
| Hand | 72.93% | 72.54% | 67.50% |
| Skeletal + Hand | 78.83% | 78.90% | 74.54% |
| Gaze + Skeletal + Hand | 82.29% | 82.10% | 82.21% |
| **Fusion** | **92.42%** | **92.37%** | **91.09%** |

The experimental results show that unimodal methods deliver poor overall performance in intention recognition. The scene modality achieves the lowest results with three metrics of 37.99%, 31.78% and 25.12%, and the gaze modality also obtains low recognition accuracy. By comparison, the skeletal motion and hand motion modalities perform better, with accuracy of 75.51% and 72.93% respectively. The recognition performance keeps improving as more modalities are integrated. The combination of skeletal and hand motion outperforms all other two-modal schemes, with its accuracy 3% higher than that of the best unimodal method. The three-modal scheme integrating gaze, skeletal and hand motion achieves the best performance among its counterparts, raising the accuracy by another 7%. The proposed multimodal fusion method obtains the optimal comprehensive performance, with three metrics reaching 93.42%, 93.37% and 92.09%. Its accuracy is improved by 18%, and it also outperforms all other approaches in F1 score and Kappa coefficient. In conclusion, the above experimental results fully demonstrate that the multimodal fusion strategy can exploit the complementary strengths of different sensing modalities and greatly improve the accuracy of human intention understanding in human-robot collaboration.

*2) Confidence-Based Multimodal Fusion Result:* Confidence of different modalities exhibit significant dynamics when the same action is performed by different participants, as shown in Fig. 12.

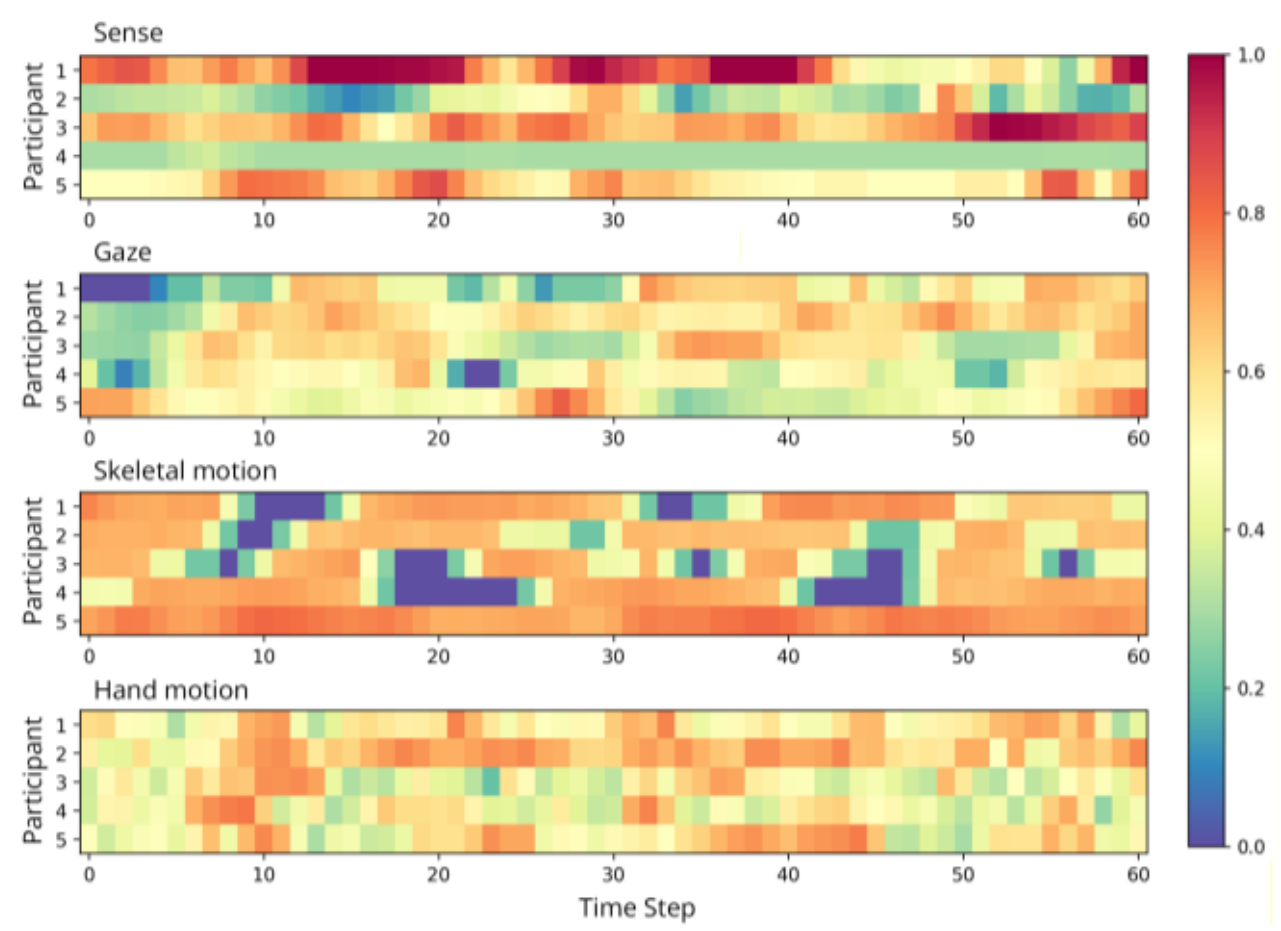

**Fig. 12.** Confidence of Different Modalities Across Participants.

Confidence heatmaps indicate that modality confidence varies greatly across participants and time steps. The sense modality shows large individual differences, the gaze modality delivers low and unstable confidence for some participants, and the skeletal motion modality suffers from severe fluctuations. By comparison, the hand motion modality maintains relatively stable performance. Such discrepancies verify the rationality of adopting confidence-aware strategies to adaptively adjust fusion weights. We further perform experiments to evaluate the effectiveness of key confidence-related modules in CAMF. We adopt Accuracy, F1 score and Standard Deviation (STDEV) as evaluation metrics, the results are shown in Table II.

TABLE II
THE RESULTS OF DIFFERENT CONFIDENCE-BASED FUSION SCHEMES

| Methods | Accuracy | F1 score | STDEV |
|---|---|---|---|
| Baseline | 90.09% | 89.83% | 0.0427 |
| w/o FC Confidence Weighting | 85.96% | 85.72% | 0.0509 |
| w/o BiLSTM Confidence Trend | 90.30% | 90.52% | 0.0340 |
| CAMF (Ours) | 91.86% | 91.84% | 0.0182 |

Experimental results show that the baseline adopts standard FC and BiLSTM networks without confidence mechanisms, with an accuracy of 90.09%, an F1 score of 89.83% and a standard deviation of 0.0427. Our CAMF model achieves the best performance, with 1.7% higher accuracy, 2.0% higher F1 score and a 57.3% lower standard deviation than the baseline. Removing the BiLSTM confidence trend module leads to a 1.56% accuracy drop and a 0.0158 rise in standard deviation. The recognition performance declines. Removing the FC confidence weighting module causes a 5.9% accuracy decline, and this variant performs worse than both the CAMF and the baseline. It proves that the BiLSTM confidence trend module alone cannot maintain performance and even underperforms the baseline, so the two modules must cooperate effectively. To evaluate robustness against noise, we compare accuracy across different noisy scenarios in Fig. 13.

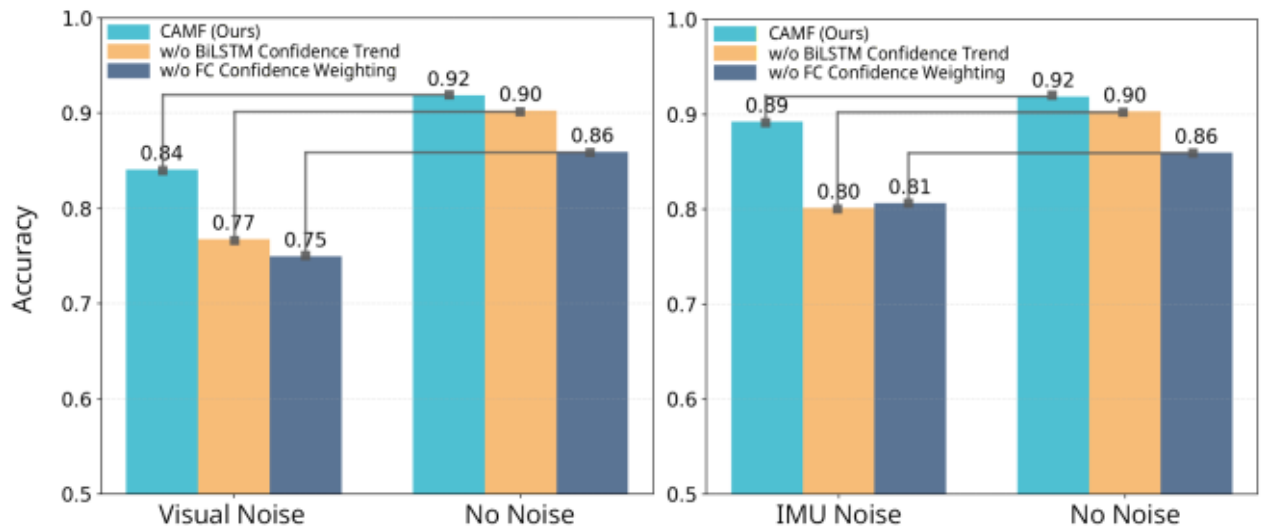

**Fig. 13.** Accuracy under Different Noise Scenarios

Under both visual and IMU noise conditions, CAMF outperforms its two ablated variants with clear margins. Compared to the noiseless setting, CAMF's accuracy drops by only 8% under visual noise and 3% under IMU noise. In contrast, removing the BiLSTM confidence trend leads to accuracy drops of 13% under visual noise and 10% under IMU noise. Removing the FC confidence weighting results in 11% and 5% drops. These results confirm that both confidence-aware modules are critical to maintaining robustness against noisy inputs.

*3) Balanced Multimodal Learning Mechanism:* The balanced multimodal learning mechanism introduces scaling weights to adaptively adjust gradient scales and dynamically regulate the contribution of each modality during training, so as to alleviate performance bias across modalities. We conduct experiments with different scaling weight configurations. The prediction scores of individual modalities are presented in Fig. 14. And we evaluate model performance in terms of accuracy, mean average precision (mAP), mean F1 score and F1 standard deviation, the results are summarized in Table III.

The scaling weight influences the learning dynamics of each modality. When $\kappa = 0$ without modality balancing, the gaze modality dominates the training process, while the skeletal motion modality and hand motion modality are suppressed, leading to severe cross-modal performance imbalance. As the scaling weight increases to the optimal setting, the prediction scores of all modalities tend to become balanced. Weak modalities receive sufficient optimization, while strong modalities no longer overly dominate the learning process.

The optimal scaling weight yields the best performance with accuracy 91.72%, mAP 97.50%, mean F1 0.9172, and F1 std 0.0136. This combination indicates high recognition precision and stable performance across categories. In contrast, inappropriate scaling leads to lower accuracy, reduced mAP, and higher F1 std, reflecting weaker precision and poorer consistency. These results confirm that the balanced multimodal learning mechanism effectively mitigates cross-modal bias for more robust fusion.

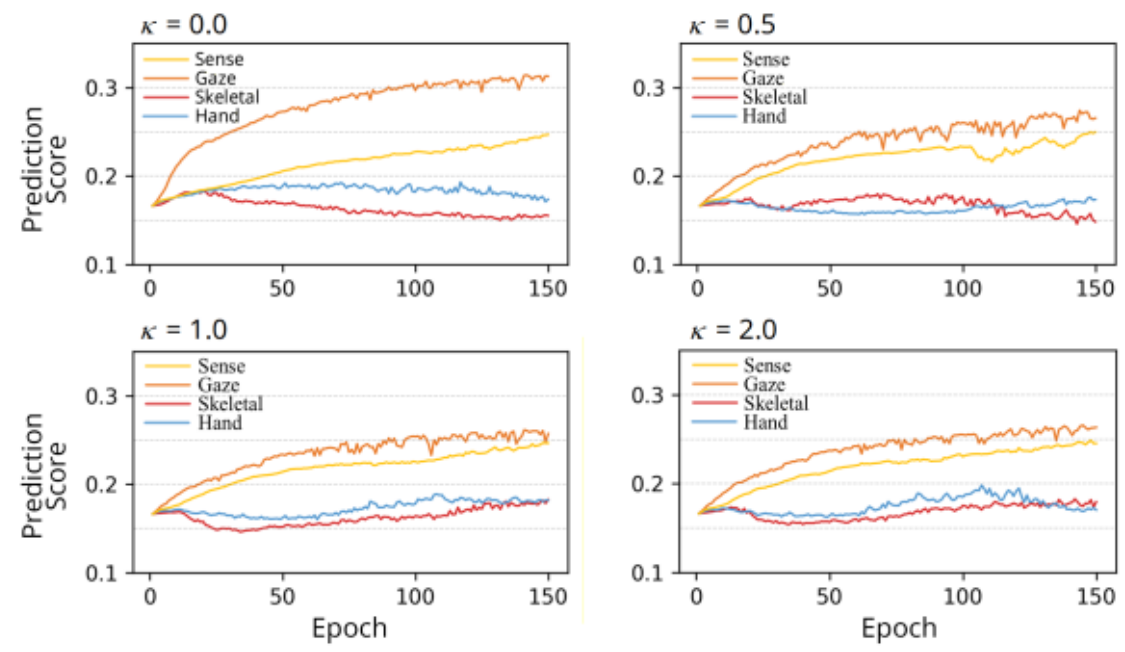

**Fig. 14.** Modality Prediction Scores under Different Scaling Weights

TABLE III

THE RESULTS OF DIFFERENT SCALING WEIGHTS

| Scaling Weight | Accuracy | mAP | F1 mean | F1 std |
|---|---|---|---|---|
| $\kappa = 0.0$ | 91.52% | 97.26% | 0.9156 | 0.0209 |
| $\kappa = 0.5$ | 91.66% | 97.18% | 0.9165 | 0.0141 |
| $\kappa = 1.0$ | 91.72% | 97.50% | 0.9172 | 0.0136 |
| $\kappa = 2.0$ | 90.71% | 96.99% | 0.9072 | 0.0212 |

### *C. Method Comparison Results*

In order to validate the effectiveness of our method, we compare the proposed CAMF with a range of mainstream and advanced multimodal fusion methods, including Dempster-Shafer evidence theory (DS), Independent Opinion Pool (IOP), Multimodal Adaptation Gate based Transformer (Transformer), Evidential Conflict-guided Late Fusion (ECoLaF), Hybrid Recurrent Neural Network (HRNN), Two-Step Long Short-Term Memory (TSLSTM), and Prototypical Modal Rebalance (PMR). The comparison results are presented in Table IV.

TABLE IV

COMPARISON RESULTS OF DIFFERENT METHODS

| Methods | Accuracy | F1 score | STDEV |
|---|---|---|---|
| DS [39] | 76.19% | 75.52% | 0.0722 |
| IOP [18] | 87.92% | 87.61% | 0.0293 |
| Transformer [40] | 80.94% | 80.41% | 0.0506 |
| ECoLaF [41] | 81.34% | 81.13% | 0.0503 |
| HRNN [42] | 86.57% | 86.21% | 0.0348 |
| TSLSTM [43] | 87.18% | 86.93% | 0.0407 |
| PMR [12] | 87.52% | 87.52% | 0.0322 |
| CAMF (Ours) | 91.86% | 91.84% | 0.0126 |

DS achieves the worst overall performance. It obtains an accuracy of 76.19%, an F1 score of 75.52%, and a high standard deviation of 0.0722. This method considers the collaboration and conflict relationships between modalities. However, it adopts fixed fusion rules. It cannot dynamically evaluate data quality and is easily disturbed by complex multimodal inputs, leading to low recognition accuracy. Transformer and ECoLaF achieve moderate performance, with accuracies of 80.94% and 81.34%, respectively. Both methods yield standard deviations higher than 0.05. Transformer integrates a multimodal adaptation gate to adjust internal model representations dynamically. Still, it lacks in-depth cross-modal feature learning and shows limited task adaptability. ECoLaF can correct model outputs according to modal conflicts. Nevertheless, it is a decision-level late fusion method and fails to extract underlying cross-modal correlation features. IOP, HRNN, TSLSTM and PMR achieve competitive results, with accuracies ranging from 86.57% to 87.92%. IOP fuses multimodal classification probabilities through independent opinion pool and Bayesian strategies to reduce prediction uncertainty. It only implements simple fusion at the result level. HRNN adopts a hybrid recurrent network to extract temporal and shallow cross-modal features effectively. However, it treats all modalities equally and ignores real-time modal confidence evaluation. TSLSTM performs multi-step intention prediction and has advantages in temporal modeling. It still lacks a dynamic modal adjustment mechanism. PMR relieves the modality imbalance problem in multimodal learning. It only optimizes the training stage and cannot adapt fusion strategies according to real-time data quality during inference. The proposed CAMF achieves the optimal performance among all methods. It reaches an accuracy of 91.86% and an F1 score of 91.84%, with a low standard deviation of 0.0126. Benefiting from the confidence-aware mechanism, CAMF dynamically evaluates modal reliability and optimizes fusion weights adaptively. It fully explores the complementary properties of multimodal data and achieves superior recognition performance.

To further evaluate the generalization capability of the proposed method, data from five participants are split for experiments. Data of four participants are used for training, and data of the remaining one for testing. Performance across different workpiece categories is shown in Fig. 15. Results under various video frame ratios are given in Fig. 16.

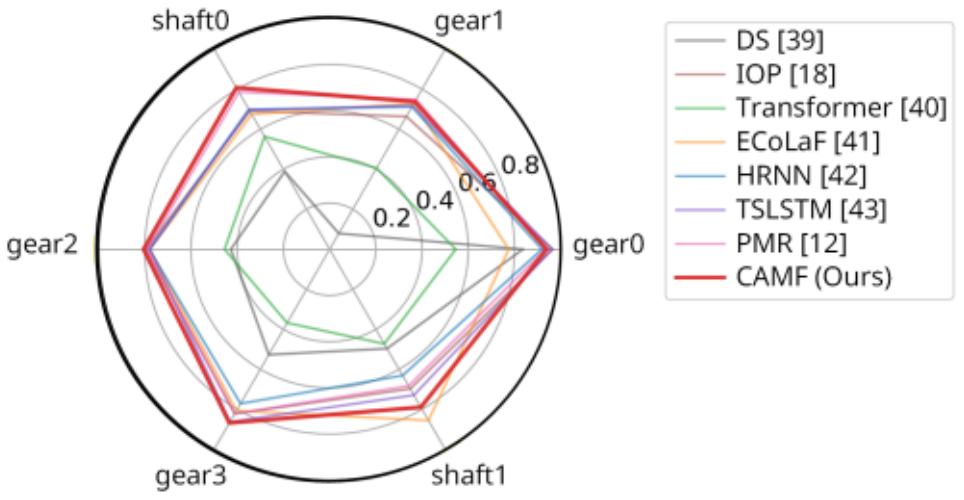

**Fig. 15.** Accuracy Comparison Across Different Workpiece Categories.

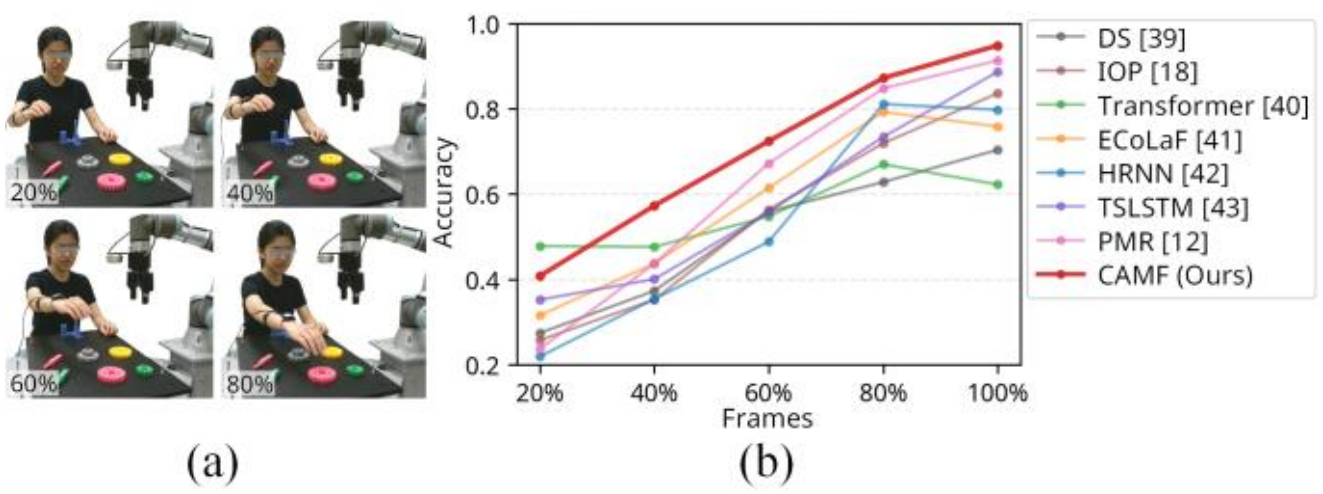

**Fig. 16.** The Results of Different Video Frame Ratios. (a) Actions under different video frame ratios. (b) Accuracy comparison.

CAMF achieves superior overall performance across all workpiece categories, maintaining consistently high and balanced recognition accuracy for both gear and shaft tasks. Its accuracy exceeds 73% in most categories, peaking at 93%. In contrast, DS exhibits significant performance fluctuations, with accuracy varying widely between 8% and 83%, showing severe deficiencies on certain workpiece types. Transformer performs poorly overall, while ECoLaF delivers the most stable results but at a generally low accuracy level. The remaining methods show performance comparable to CAMF but remain consistently inferior across the board. At different video frame ratios, CAMF achieves competitive accuracy across all settings, with values of 35%, 52%, 67%, 82% and 89%. Its performance rises steadily as the number of valid frames increases. Transformer has low overall accuracy yet performs well at low frame ratios, as it relies on basic visual features and makes full use of shallow information from limited frames. PMR shows the fastest accuracy growth, indicating its strong capability to extract temporal features from consecutive frames. The results demonstrate that the proposed method delivers better overall performance and stronger generalization across diverse workpieces and varying input conditions.

*D. Case Study*

To verify the practicability and adaptability of CAMF in realistic complex workshop scenarios, we conduct two groups of industrial case experiments in this subsection. We first simulate common interferences such as low light and human occlusion to evaluate the model's environmental adaptability. Then we implement a complete human-robot collaborative assembly workflow to validate the performance of the method in real-time intention recognition and robotic cooperative tasks.

This set of tests adopts the pre-trained model, with no supplementary training on occlusion datasets. We establish four typical interference conditions including low light, upper body occlusion, face occlusion and full body occlusion, as illustrated in Fig. 17. Specifically, the low light environment is generated by directly lowering the brightness of original RGB images, and the corresponding quantitative results under various interferences are summarized in Table V.

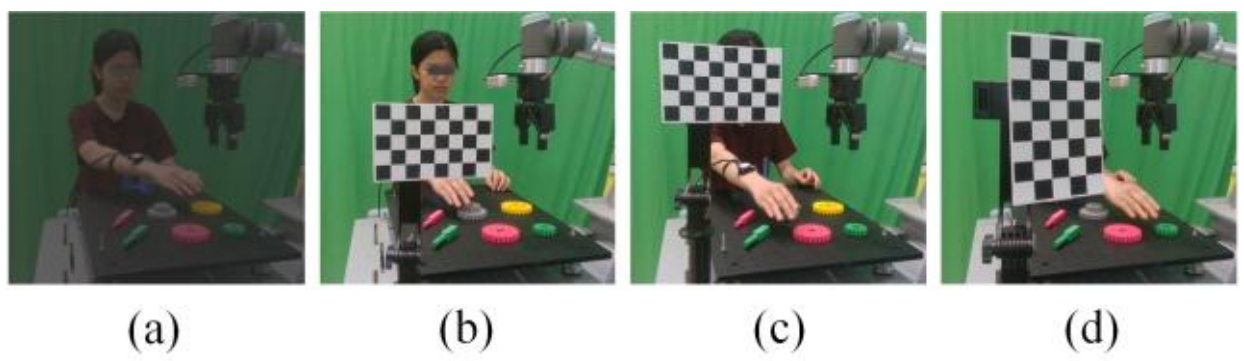

**Fig. 17.** Performance test under different interference conditions. (a) Low light scenario. (b) Upper body occlusion. (c) Face occlusion. (d) Full body occlusion.

TABLE V
THE RESULTS OF DIFFERENT INTERFERENCE CONDITIONS

| Interference Scenario | Accuracy | F1 score | STDEV |
|---|---|---|---|
| No interference | 87.86% | 88.03% | 0.0435 |
| Low light scenario | 83.58% | 83.76% | 0.0535 |
| Upper body occlusion | 43.17% | 36.49% | 0.2994 |
| Face occlusion | 77.14% | 77.73% | 0.1035 |
| Full body occlusion | 30.00% | 20.15% | 0.3274 |

Experimental results indicate that external interferences impair recognition performance to different extents. The model achieves an accuracy of 87.86%, an F1 score of 88.03% and a standard deviation of 0.0435 under normal conditions. The accuracy drops by 4% in the low light environment. It maintains acceptable performance under face occlusion with an 11% accuracy decline, which proves that the proposed method can handle partial occlusion effectively. The performance drops sharply to an accuracy of 43.17% under upper body occlusion, and full body occlusion produces the lowest accuracy at 30.00%. Effective human features are almost completely lost under such extreme occlusion, and this is also a challenge for all visual recognition methods.

We conduct a complete human-robot collaborative assembly test to evaluate the application of the proposed CAMF method in real-world task scenarios. The full process includes user intention expression, intention recognition, robotic grasping and delivery. The sequential stages of the task are illustrated in Fig. 18.

The sequential snapshots show the complete collaborative manipulation process. Steps (1) to (4) record the user's intention expression, and the system recognizes the target workpiece at Step (4). Steps (5) and (6) present the grasping phase: the system uses YOLOv11 with a custom-built dataset for target detection, extracts the target mask, and adopts GraspNet to generate grasping schemes. It selects the optimal grasp pose under UR3 workspace constraints to realize stable picking. Steps (7) to (11) show workpiece delivery, where the UR3 moves along the safe gate trajectory to prevent collisions. The robot returns to its home position at Step (12) to end the task. In line with prior tests on video frame ratios, the robot starts working when the user's motion reaches 60% to 80%, enabling proactive collaboration.

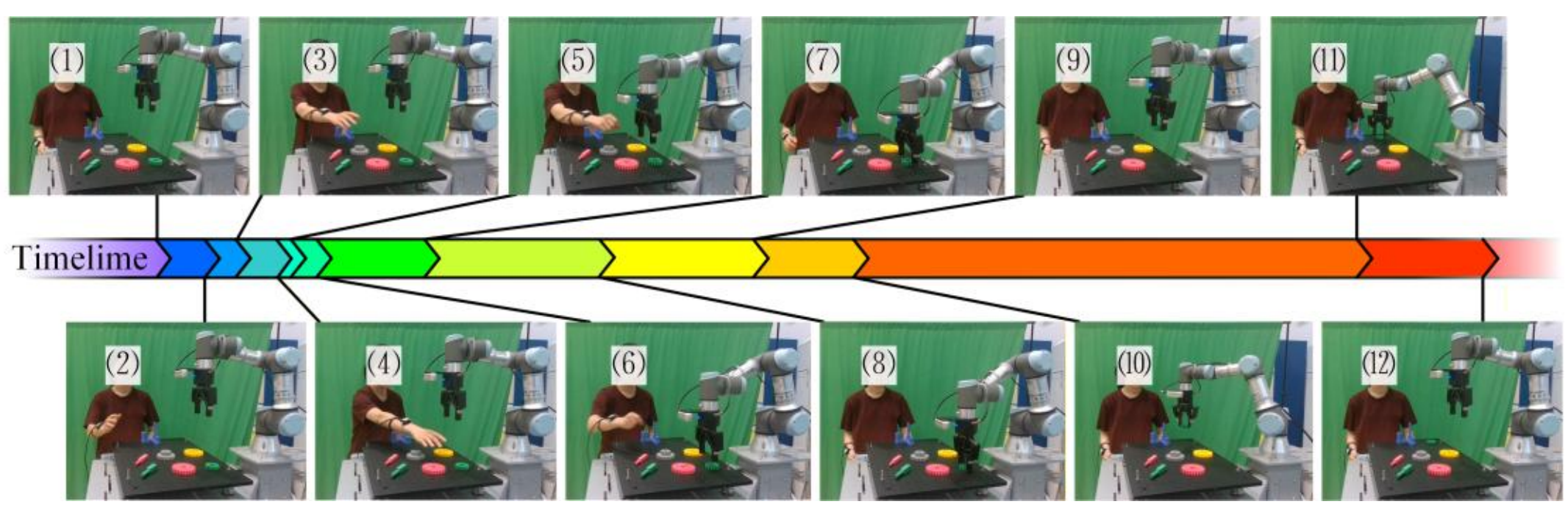

**Fig. 18.** Human-Robot Collaborative Manipulation Process Timeline

## IV. DISCUSSION

### *A. Results Discussion*

Experimental results show that the proposed CAMF achieves excellent performance and environmental adaptability in human-robot intention recognition. It combines multiple modalities to make up for the insufficient feature information of single modalities and improves recognition accuracy. Its confidence trend-driven fusion and multimodal balanced learning assess modal reliability in real time, adjust fusion weights and training gradients, and reduce the impact of low-quality data. Compared with other mainstream methods, CAMF delivers higher accuracy and smaller deviation, and runs stably on different workpieces and motion ratios. It also maintains reliable performance under mild interference, which adapts well to complex industrial environments. However, this method still has limitations. Extreme upper-body or full-body occlusion causes massive loss of valid features and a sharp drop in modal confidence, leading to severe performance degradation. Moreover, the balanced learning mechanism only optimizes modal bias during training. It fails to respond rapidly to sudden modal failure in extreme scenarios and may result in occasional recognition errors.

### *B. System Evaluation*

The constructed human-robot collaboration system exhibits good practicability and industrial adaptability. It integrates visual scene, gaze, skeletal motion, and hand inertial sensing data to comprehensively capture human collaborative intentions and support high-precision active human-robot collaboration. The hardware platform adopts depth cameras and IMU sensors for synchronous multimodal acquisition, and cooperates with the UR3 robotic arm and adaptive gripper to steadily complete target detection, workpiece grasping, and safe delivery. The software framework based on ROS2 realizes real-time data transmission among sensors, workstations and robots, and improves the fluency and efficiency of human-robot cooperation through the adaptive confidence fusion strategy. Nevertheless, this system still has engineering limitations. It is only validated on standard assembly parts with relatively single scenarios, lacking sufficient verification in tasks with special-shaped workpieces, dynamic disturbances and multi-stage processes, which limits its overall application scope. Moreover, the current model only supports offline training and fixed-parameter inference without online adaptive optimization, making it difficult to adapt to variable operating habits and dynamic industrial environments.

### C. Future Work

To address the limitations of this study, we will optimize the algorithm and add feature compensation functions to improve recognition performance and response speed under strong interference. We will also equip the system with online learning to adjust model parameters in real time for different users and changing working conditions. Besides, we will apply the multimodal fusion method to robot motion control and interface interaction tasks to expand its application potential and promote practical deployment in industrial scenarios.

## V. Conclusion

A confidence-aware multimodal fusion framework (CAMF) is developed to facilitate intelligent human-robot collaboration in industrial assembly scenarios. This work combines multiple types of sensing information to comprehensively perceive human behaviors, and adopts BiLSTM to model temporal dependencies within continuous operation sequences. Targeting the dynamic changes and noise disturbance of different data sources, confidence assessment modules are introduced to conduct adaptive feature fusion. Meanwhile, a set of optimized learning strategies are designed to balance the contribution of each modality during training and restrain the adverse impact of unreliable data.

The proposed framework organically integrates dynamic fusion rules and improved training mechanisms. It can adapt to varying working conditions and mitigate the defects caused by static fusion schemes and imbalanced multimodal learning. In real-world collaborative operations, the system is able to perceive human intentions and coordinate robotic actions accordingly. It exhibits good environmental adaptability and steady running performance when confronted with common industrial disturbances. The overall performance validates that this solution is well suited for practical industrial human-robot interaction tasks, and provides a feasible reference for the design of robust multimodal perception systems in similar application scenarios.

## References

[1] L. Pérez, S. Rodríguez-Jiménez, N. Rodríguez, R. Usamentiaga, D. F. García, and L. Wang, "Symbiotic human–robot collaborative approach for increased productivity and enhanced safety in the aerospace manufacturing industry," *Int. J. Adv. Manuf. Technol.*, vol. 106, no. 3, pp. 851–863, Jan. 2020.

[2] Q. Wang, H. Liu, F. Ore, L. Wang, J. B. Hauge, and S. Meijer, "Multi-actor perspectives on human robotic collaboration implementation in the heavy automotive manufacturing industry—A Swedish case study," *Technol. Soc.*, vol. 72, Art. no. 102165, Oct. 2023.

[3] M. Marinelli, "From Industry 4.0 to Construction 5.0: Exploring the path towards human–robot collaboration in construction," *Systems*, vol. 11, no. 3, Art. no. 152, Mar. 2023.

[4] A. Baratta, A. Cimino, M. G. Gnoni, and F. Longo, "Human robot collaboration in industry 4.0: a literature review," *Procedia Comput. Sci.*, vol. 217, pp. 1887–1895, 2023.

[5] P. Zheng, S. Li, J. Fan, C. Li, and L. Wang, "A collaborative intelligence-based approach for handling human-robot collaboration uncertainties," *CIRP Ann.*, vol. 72, no. 1, pp. 1–4, 2023.

[6] A. K. Inkulu, M. R. Bahubalendruni, A. Dara, and S. K., "Challenges and opportunities in human robot collaboration context of Industry 4.0—a state of the art review," *Ind. Robot*, vol. 49, no. 2, pp. 226–239, 2022.

[7] S. S. Kumar, S. R. Kumar, and G. Ramesh, "From Industry 4.0 to 5.0: Enriching manufacturing excellence through Human–Robot interaction and technological empowerment," in *Intelligent Systems and Industrial Internet of Things for Sustainable Development*, Chapman and Hall/CRC, 2024, pp. 24–51.

[8] H. Su, W. Qi, J. Chen, C. Yang, J. Sandoval, and M. A. Laribi, "Recent advancements in multimodal human–robot interaction," *Front. Neurorobot.*, vol. 17, Art. no. 1084000, 2023.

[9] T. Wang, P. Zheng, S. Li, and L. Wang, "Multimodal human–robot interaction for human-centric smart manufacturing: a survey," *Adv. Intell. Syst.*, vol. 6, no. 3, Art. no. 2300359, 2024.

[10] T. Xue, W. Wang, J. Ma, W. Liu, Z. Pan, and M. Han, "Progress and prospects of multimodal fusion methods in physical human–robot interaction: A review," *IEEE Sens. J.*, vol. 20, no. 18, pp. 10355–10370, 2020.

[11] X. Peng, Y. Wei, A. Deng, D. Wang, and D. Hu, "Balanced multimodal learning via on-the-fly gradient modulation," in *Proc. IEEE/CVF Conf. Comput. Vis. Pattern Recognit.*, 2022, pp. 8238–8247.

[12] Y. Fan, W. Xu, H. Wang, J. Wang, and S. Guo, "PMR: Prototypical modal rebalance for multimodal learning," in *Proc. IEEE/CVF Conf. Comput. Vis. Pattern Recognit.*, 2023, pp. 20029–20038.

[13] H. Liu, T. Fang, T. Zhou, and L. Wang, "Towards robust human-robot collaborative manufacturing: Multimodal fusion," *IEEE Access*, vol. 6, pp. 74762–74771, 2018.

[14] S. Liu, L. Wang, and X. V. Wang, "Function block-based multimodal control for symbiotic human–robot collaborative assembly," *J. Manuf. Sci. Eng.*, vol. 143, no. 9, pp. 091001, Sep. 2021.

[15] J. Duan, Y. Fang, Q. Zhang, and J. Qin, "HRC for dual-robot intelligent assembly system based on multimodal perception," in *Proc. Inst. Mech. Eng. Part B J. Eng. Manuf.*, vol. 238, no. 4, pp. 562–576, 2024.

[16] S. Rafael, "The contribution of early multimodal data fusion for subjectivity in HCI," in *2021 16th Iberian Conf. Inf. Syst. Technol. (CISTI)*, 2021, pp. 1–6.

[17] X. Luo et al., "Noise-augmented multi-modal entity alignment with confidence-based dynamic fusion," *Sci. China Inf. Sci.*, to be published, 2025.

[18] S. Trick, D. Koert, J. Peters, and C. A. Rothkopf, "Multimodal uncertainty reduction for intention recognition in human-robot interaction," in *2019 IEEE/RSJ Int. Conf. Intell. Robots Syst. (IROS)*, 2019, pp. 7009–7016.

[19] S. Jiang et al., "A novel human-in-the-loop multimodal intention fusion method for human-robot interaction," *IEEE Trans. Autom. Sci. Eng.*, to be published, 2025.

[20] Y. Yang, Y. Q. Yang, G. Ren, and B. G. Yu, "Hierarchically trusted evidential fusion method with consistency learning for multimodal language understanding," *Knowl.-Based Syst.*, vol. 312, Art. no. 113164, 2025.

[21] Z. Han, F. Yang, J. Huang, C. Zhang, and J. Yao, "Multimodal dynamics: Dynamical fusion for trustworthy multimodal classification," in *Proc. IEEE/CVF Conf. Comput. Vis. Pattern Recognit.*, 2022, pp. 20707–20717.

[22] X. Zhao et al., "Learning multimodal confidence for intention recognition in human-robot interaction," *IEEE Robot. Autom. Lett.*, to be published, 2024.

[23] J. Martinez, M. J. Black, and J. Romero, "On human motion prediction using recurrent neural networks," in *Proc. IEEE Conf. Comput. Vis. Pattern Recognit. (CVPR)*, 2017, pp. 4674–4683.

[24] E. Aksan, M. Kaufmann, and O. Hilliges, "Structured prediction helps 3D human motion modelling," in *Proc. IEEE/CVF Int. Conf. Comput. Vis. (ICCV)*, Oct. 2019, pp. 7144–7153.

[25] X. Liu, J. Yin, J. Liu, P. Ding, J. Liu, and H. Liu, "TrajectoryCNN: A new spatio-temporal feature learning network for human motion prediction," *IEEE Trans. Circuits Syst. Video Technol.*, vol. 31, no. 6, pp. 2133–2146, Jun. 2021.

[26]C. Zhong, L. Hu, Z. Zhang, Y. Ye, and S. Xia, “Spatio-temporal gating adjacency GCN for human motion prediction,” in *Proc. IEEE/CVF Conf. Comput. Vis. Pattern Recognit. (CVPR)*, 2022, pp. 6437–6446.

[27]E. Barsoum, J. Kender, and Z. Liu, “HP-GAN: Probabilistic 3D human motion prediction via GAN,” in *Proc. IEEE/CVF Conf. Comput. Vis. Pattern Recognit. Workshops (CVPRW)*, Jun. 2018, pp. 1418–1427

[28]S. Aliakbarian, F. Sadat Saleh, M. Salzmann, L. Petersson, and S. Gould, “A stochastic conditioning scheme for diverse human motion prediction,” in *Proc. IEEE/CVF Conf. Comput. Vis. Pattern Recognit. (CVPR)*, 2020, pp. 5222–5231.

[29]E. Aksan, M. Kaufmann, P. Cao, and O. Hilliges, “A spatio-temporal transformer for 3D human motion prediction,” in *Proc. Int. Conf. 3D Vis. (3DV)*, Dec. 2021, pp. 565–574.

[30]L. Chen, R. Liu, W. Zhang, Y. Hou, Q. Zhang, and D. Zhou, “MSTP-Net: Multiscale spatio-temporal parallel networks for human motion prediction,” *IEEE Trans. Consum. Electron.*, vol. 70, no. 1, pp. 3318–3331, Feb. 2023.

[31]S. Hochreiter and J. Schmidhuber, “Long short-term memory,” *Neural Comput.*, vol. 9, no. 8, pp. 1735–1780, Nov. 1997.

[32]R. Chellali, “Predicting arm movements: A multi-variate LSTM-based approach for human-robot hand clapping games,” in *2018 27th IEEE Int. Symp. Robot Hum. Interact. Commun. (RO-MAN)*, Aug. 2018, pp. 1137–1142.

[33]Z. Liu, Q. Liu, W. Xu, Z. Liu, Z. Zhou, and J. Chen, “Deep learning-based human motion prediction considering context awareness for human-robot collaboration in manufacturing,” *Procedia CIRP*, vol. 83, pp. 272–278, 2019.

[34]C. Ma et al., “A bi-directional LSTM network for estimating continuous upper limb movement from surface electromyography,” *IEEE Robot. Autom. Lett.*, vol. 6, no. 4, pp. 7217–7224, 2021.

[35]B. Wen, W. Yang, J. Kautz and S. Birchfield, "FoundationPose: Unified 6D Pose Estimation and Tracking of Novel Objects," in *Proc. IEEE Conf. Comput. Vis. Pattern Recognit. (CVPR)*, 2024, pp. 17868-17879.

[36]F. Ryan, A. Bati, S. Lee, D. Bolya, J. Hoffman and J. M. Rehg, "Gaze-LLE: Gaze Target Estimation via Large-Scale Learned Encoders," in *Proc. IEEE Conf. Comput. Vis. Pattern Recognit. (CVPR)*, 2025, pp. 28874-28884.

[37]S. Yan, Y. Xiong and D. Lin, “Spatial temporal graph convolutional networks for skeleton-based action recognition,” in *Proc. AAAI conf. Artif. Intel.*, vol. 32, no. 1, Apr. 2018.

[38]S. Zhang, D. Zheng, X. Hu and M. Yang, “Bidirectional Long Short-Term Memory Networks for Relation Classification,” in *Proc. Pac. Asia Conf. Lang. Inf. Comput.*, 2015.

[39]Z. Jiao, “Research on multimodal human-computer interaction technology based on audiovisual fusion,” in *Proc. 7th Int. Conf. Intell. Comput. Signal Process. (ICSP)*, 2022, pp. 1378–1381.

[40]W. Rahman, M. K. Hasan, S. Lee, A. B. Zadeh, C. Mao, L. P. Morency, and E. Hoque, “Integrating multimodal information in large pretrained transformers,” in *Proc. 58th Annu. Meet. Assoc. Comput. Linguist. (ACL)*, Jul. 2020, pp. 2359–2369.

[41]L. Deregnaucourt, H. Laghmara, A. Lechervy, and S. Ainouz, “A Conflict-Guided Evidential Multimodal Fusion for Semantic Segmentation,” in *Proc. IEEE/CVF Winter Conf. Appl. Comput. Vis. (WACV)*, Feb. 2025, pp. 1373–1382.

[42]X. Gao, L. Yan, G. Wang, and C. Gerada, “Hybrid recurrent neural network architecture-based intention recognition for human–robot collaboration,” *IEEE Trans. Cybern.*, vol. 53, no. 3, pp. 1578–1586, 2021.

[43]P. U. Ravva, P. Kullu, M. F. Abrar, and R. L. Barmaki, “A Machine Learning Approach for Predicting Upper Limb Motion Intentions with Multimodal Data in Virtual Reality,” arXiv preprint arXiv:2405.13023, 2024.